\documentclass{article}

\usepackage[preprint]{corl_2026} 

\usepackage{graphicx}
\usepackage{algorithm}
\usepackage{algpseudocode}
\usepackage{wrapfig}
\usepackage{booktabs}
\usepackage{multirow}
\usepackage[normalem]{ulem}
\usepackage[dvipsnames]{xcolor}
\usepackage{tabularx}

\definecolor{simColor}{HTML}{d63384}
\definecolor{pColor}{HTML}{7c3aed}
\definecolor{lColor}{HTML}{0891b2}
\definecolor{eColor}{HTML}{ea580c}

\usepackage{xspace}
\newcommand{\ours}{SIMPLE\xspace}
\newcommand{\ntask}{16\xspace}

\newcommand{\thickrule}{\noalign{\hrule height 0.9pt}}
\newcommand{\thinrule}{\noalign{\hrule height 0.4pt}}
\newcommand{\tablesp}{\noalign{\vskip 2pt}}
\newcommand{\ul}[1]{\uline{#1}}

\title{\ours: Simulation-Based Policy Learning and Evaluation for Humanoid Loco-manipulation}

\author{
  Songlin Wei\textsuperscript{$*$}, 
  Zhenhao Ni\textsuperscript{$*$}, 
  Jie Liu\textsuperscript{$*$}, 
  Zhenyu Zhao\textsuperscript{$*$}, \\
  Junjie Ye,
  Hongyi Jing, 
  Junkai Xia,
  Xiawei Liu,
  Michael Leong, \\
  Liang Heng, 
  Di Huang,
  Yue Wang$^\dag$ \\ 
  USC Physical Superintelligence (PSI) Lab \\
  {$*$} equal contribution, $^\dag$ corresponding author \\{\hypersetup{urlcolor=simColor}\url{https://psi-lab.ai/SIMPLE}}
}
\begin{document}
\maketitle

\vspace{-2em}
\begin{figure}[h]
\centering
\includegraphics[width=0.98\linewidth]{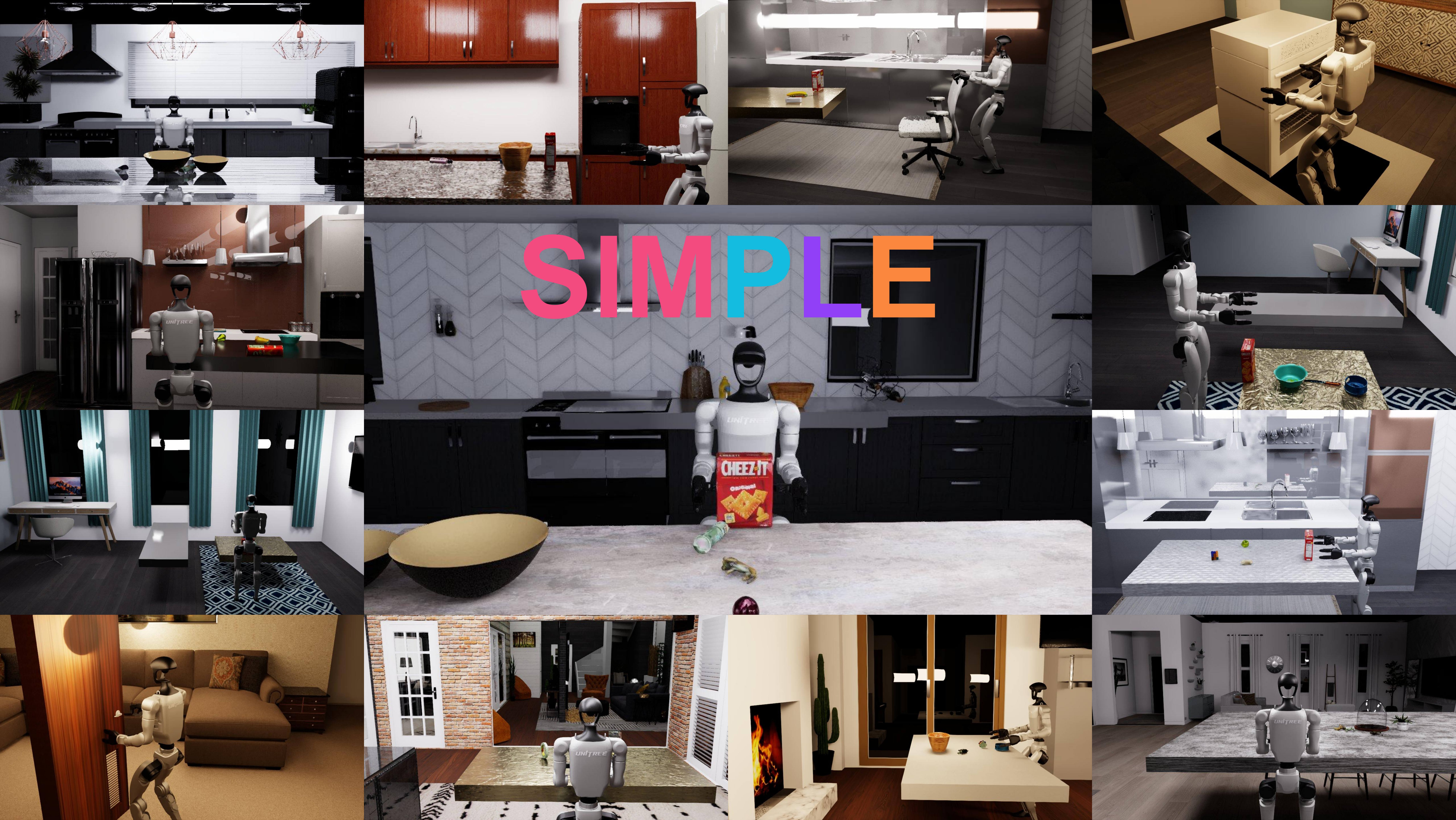}
\vspace{-2mm}
\caption{\textbf{Humanoid Loco-manipulation in \ours.} 
We introduce a comprehensive simulation benchmark designed to standardize the evaluation and training of humanoid foundation models.
The framework features \ntask diverse whole-body tasks across 50 indoor scenes, utilizing over 1,000 object assets.
By coupling MuJoCo's robust contact physics with Isaac Sim's photorealistic rendering, \ours provides built-in data collection pipelines and natively benchmarks state-of-the-art VLAs and WAMs.
}
\vspace{-4mm}
\label{fig:teaser}
\end{figure}

\begin{abstract}
Humanoid foundation models are advancing faster than we can evaluate them.
While real-world testing is expensive and difficult to reproduce, existing simulation benchmarks focus primarily on table-top or wheeled robots.
A scalable and reproducible benchmark for whole-body humanoid loco-manipulation remains an open problem.
To this end, we present \ours, a unified simulation testbed for humanoid policy learning and evaluation.
\ours couples the accurate contact-rich dynamics of MuJoCo with the photorealistic rendering of IsaacSim.
It provides a large-scale environment comprising \ntask diverse whole-body tasks, 50 indoor scenes, and over 1,000 object assets.
To facilitate scalable data collection, the framework integrates two data generation pipelines: automated trajectory generation via motion planning and a low-latency VR teleoperation interface.
We further integrate and benchmark mainstream humanoid policies at scale in \ours, including lightweight imitation networks, large vision-language-action (VLA) models, and recent world action models (WAMs).
Our experiments demonstrate that policies trained on data collected in \ours can be transferred zero-shot to physical humanoid robots under similar settings, providing a robust and reproducible foundation for humanoid robotics research. 
\end{abstract}


\section{Introduction}

Robotic foundation models~\cite{bjorck2025gr00t,dreamzero,intelligencepi05,wei2026psi0} are advancing rapidly, demonstrating increasingly general behaviors.
However, scaling these models is fundamentally constrained by evaluation.
For table-top and wheeled robots, standardized simulation benchmarks~\cite{chen2025robotwin,geng2025roboverse,liu2023libero,li2023behavior} have greatly accelerated progress.
Developing equivalent benchmarks for humanoid robots remains an open challenge.
Humanoid loco-manipulation is inherently complex, requiring the tight coordination of dynamic locomotion, whole-body balancing, and fine-grained dexterous manipulation.

Currently, most humanoid foundation models are evaluated directly in the real world.
While authentic, real-world evaluation is expensive, time-consuming, and notoriously difficult to reproduce.
Consequently, conducting fair, side-by-side comparisons across different policies is a major challenge.
Recent efforts to automate real-world evaluation via crowd-sourced rankings \cite{atreya2025roboarena} or VLM-based reset policies \cite{zhou2025autoeval} face scalability limits and potential biases.
Simulation offers a scalable alternative \cite{li2024simplerenv}, but existing platforms are highly fragmented.
They either focus exclusively on tabletop manipulation~\cite{chen2025robotwin,geng2025roboverse}, ignoring locomotion and whole-body balancing, or on pure locomotion~\cite{humanoidverse,sferrazza2024humanoidbench}, lacking diverse object interactions.

To address this fragmentation, we present \textbf{\ours}, a comprehensive, full-stack simulation testbed designed for the extensive evaluation and fair comparison of humanoid policies.
Instead of focusing on a narrow subset of skills, \ours provides a large-scale environment dedicated to whole-body loco-manipulation.
We design \ntask diverse tasks, spanning rigid object pick-and-place, non-prehensile interaction, and articulated object manipulation.
These tasks are set in 50 indoor scenes \cite{khanna2024habitat} utilizing over 1,000 object assets \cite{deitke2023objaverse}.
To support this scale with high fidelity, \ours utilizes a hybrid architecture, coupling the accurate contact-rich dynamics of MuJoCo \cite{todorov2012mujoco} with the photorealistic ray-traced rendering of Isaac Sim \cite{NVIDIA_Isaac_Sim}.
This dual-simulator approach ensures that policies trained in \ours experience both the rigorous physical constraints of real-world locomotion and the complex visual diversity required for robust perception.

A standardized testbed must also support the full lifecycle of policy development.
Beyond evaluation, \ours facilitates scalable data collection by providing two built-in pipelines: automated trajectory generation driven by motion planners \cite{curobo_v2,chen2024bodex}, and a low-latency VR teleoperation interface for collecting human demonstrations.
By standardizing these pipelines, \ours eliminates the engineering overhead typically required to gather high-quality, whole-body demonstration data.
Furthermore, the testbed integrates state-of-the-art lower-body controllers \cite{li2025amo,luo2025sonic} and offers a unified interface with native compatibility for mainstream vision-language-action (VLA) architectures, such as $\Psi_0$ \cite{wei2026psi0}, $\pi_{0.5}$ \cite{intelligencepi05}, and GR00T-N1.6 \cite{bjorck2025gr00t}, as well as World-Action-Models (WAMs) like DreamZero~\cite{dreamzero} and other recent models \cite{yang2025egovla, bi2026h}.
Crucially, \ours is designed to be fully portable.
This modularity enables researchers to easily extend the framework with new tasks, objects, environments, and policies.

We conduct extensive evaluations on \ours to benchmark state-of-the-art policies at scale.
Our experiments reveal a strong correlation between policy performance in simulation and in the real world.
We further demonstrate that policies trained on data collected in \ours can be transferred zero-shot to physical humanoid robots under similar settings.
Our contributions are threefold:
\vspace{-2mm}
\begin{itemize}
\setlength{\itemsep}{0pt}
\setlength{\parsep}{0pt}
    \item We introduce \ours, a hybrid full-stack simulation framework standardizing the evaluation of humanoid loco-manipulation.
    \item We construct a large-scale loco-manipulation environment comprising \ntask diverse tasks, over 1,000 objects, and 50 scenes, alongside built-in pipelines for automated and teleoperated data collection, yielding over 6,000 trajectories.
    \item We conduct extensive benchmarking of mainstream VLA and WAM policies, demonstrating strong sim-to-real correlation and effective zero-shot policy transfer.
\end{itemize}
\vspace{-2mm}

\vspace{-1mm}
\section{Related Work}
\vspace{-1mm}

\vspace{-1mm}
\subsection{Robot Policy Evaluation in the Real World}
\vspace{-1mm}
Evaluating generalist policies \cite{intelligencepi05, bjorck2025gr00t, deng2025graspvla, zhang2024uni, wei2026psi0} directly in the real world is the current standard but presents significant scalability challenges.
Fair and reproducible evaluation requires multiple rollouts with meticulous environment resets, which are difficult to standardize and prone to human-operator bias.
To address this, prior works have proposed using 3D-printable assets \cite{fang2020graspnet, wei2024d} or centralized physical testbeds \cite{yakefu2025robochallenge}, though these still require substantial human supervision.
While AutoEval \cite{zhou2025autoeval} employs large language models (LLMs) to automate resets, the robustness of LLM-based physical interventions remains suboptimal.
Alternatively, RoboArena \cite{atreya2025roboarena} mitigates human bias via a decentralized, double-blind evaluation protocol.
However, while effective for zero-shot testing on controlled tabletop setups with standardized datasets \cite{khazatsky2024droid}, this paradigm does not easily extend to humanoids.
Because large-scale humanoid data is scarce, evaluating VLA models typically requires collecting in-domain demonstrations for policy fine-tuning prior to testing.
Both collecting these demonstrations and conducting subsequent physical evaluations involve highly dynamic, full-body motions with heavy hardware, significantly increasing operational complexity and safety risks.
Consequently, while real-world evaluation remains the gold standard, relying on it for the rapid, iterative development of humanoid policies is highly impractical.

\vspace{-1mm}
\subsection{Simulation-Based Benchmarks for Manipulation}
\vspace{-1mm}
Simulation provides a scalable, reproducible, and automated alternative to real-world testing \cite{wang2025roboeval}.
Benchmarks such as RLBench \cite{james2020rlbench} and LIBERO \cite{liu2023libero} have driven significant progress in tabletop single-arm manipulation, while platforms like RoboCasa \cite{robocasa365}, Behavior-1K \cite{li2023behavior}, MolmoSpaces \cite{molmospaces2026}, and ManiSkill-HAB \cite{shukla2024maniskill} have extended evaluation to mobile manipulation.
Crucially, Simpler-Env \cite{li2024simplerenv} demonstrated that simulation can serve as a faithful proxy for real-world evaluation.
By identifying and mitigating visual and control gaps, they showed that policy performance in simulation is highly correlated with real-world rankings.
More recently, RoboTwin \cite{chen2025robotwin} introduced extensive domain randomization for bimanual tasks, and RoboVerse \cite{geng2025roboverse} unified multiple benchmarks within IsaacLab \cite{mittal2025isaaclab}.
However, these platforms are typically built on physics engines or configurations optimized exclusively for fixed-base or wheeled robots \cite{chen2025robohanger}.
Consequently, they lack the necessary physics support and integration for humanoid whole-body loco-manipulation.

\vspace{-1mm}
\subsection{Simulation for Humanoid Loco-Manipulation}
\vspace{-1mm}
Simulation has long been the primary paradigm for training robust humanoid locomotion policies before real-world deployment.
Recent frameworks like AMO \cite{li2025amo} and SONIC \cite{luo2025sonic} leverage massively parallel reinforcement learning in IsaacGym \cite{makoviychuk2021isaacgym} and IsaacLab \cite{mittal2025isaaclab} to train highly stable lower-body and whole-body tracking networks.
These locomotion-focused policies are frequently deployed and evaluated in MuJoCo \cite{todorov2012mujoco} due to its superior contact fidelity.
However, while these environments excel at locomotion, they lack the diverse object assets, complex indoor scenes, and photorealistic rendering required to train and evaluate generalist foundation models.
Conversely, platforms capable of high-fidelity rendering (such as those built on PhysX) often struggle to provide the accurate, high-frequency contact dynamics necessary for stable humanoid locomotion and delicate hand-object interaction.
\ours bridges this divide by natively building upon MuJoCo \cite{todorov2012mujoco} for superior contact fidelity, ensuring compatibility with state-of-the-art locomotion policies, while coupling it with Isaac Sim's photorealistic rendering \cite{NVIDIA_Isaac_Sim}.
To the best of our knowledge, \ours is the first comprehensive, full-stack simulation testbed dedicated to whole-body humanoid loco-manipulation, seamlessly integrating data generation, policy learning, and standardized evaluation.

\vspace{-5mm}
\section{Method}
\begin{figure}[!t]
    \centering
    \includegraphics[width=1\linewidth]{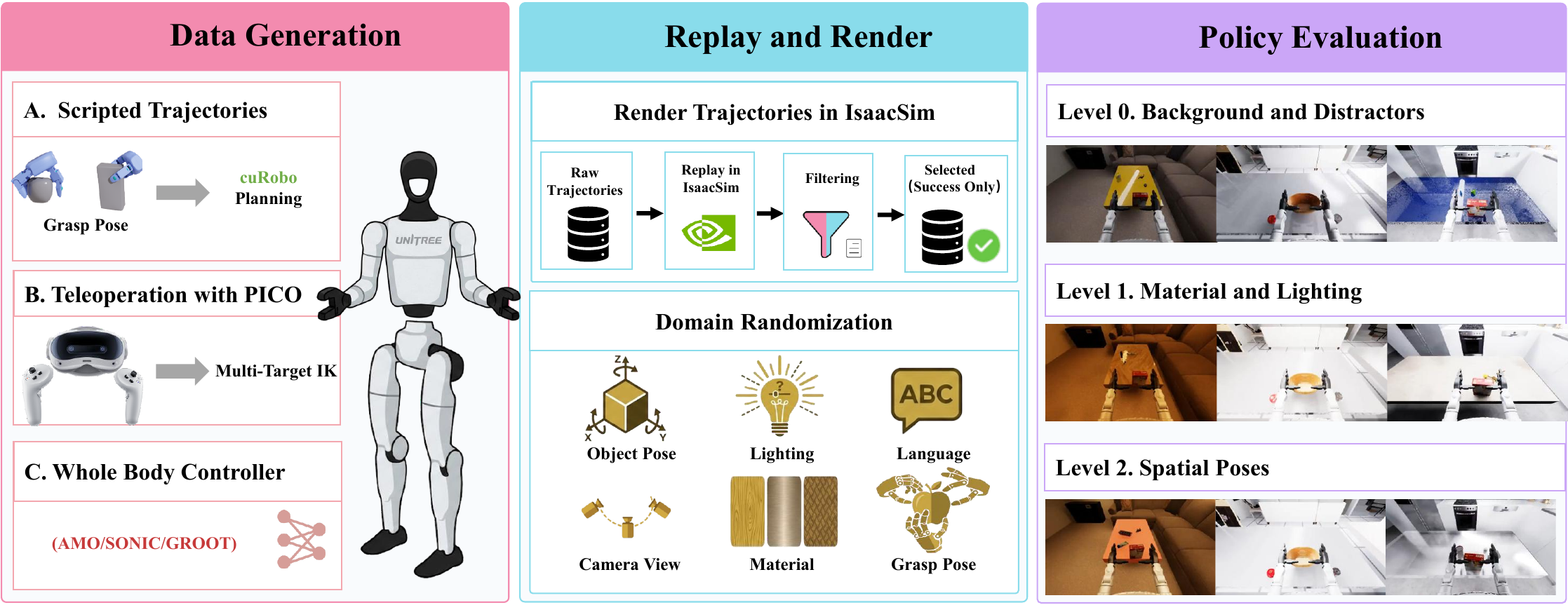}
    \vspace{-5mm}
    \caption{\textbf{System Pipeline.} Our pipeline consists of three stages: (1) data generation in MuJoCo via motion planning and teleoperation; (2) offline replay and rendering in Isaac Sim to obtain photorealistic visual observations; and (3) policy evaluation under diverse domain-randomized settings.}
    \vspace{-3mm}
    \label{fig:pipeline}
\end{figure}

\vspace{-1mm}
We present \ours, a comprehensive simulation infrastructure designed for humanoid loco-manipulation.
The framework integrates dual-simulators for physics and rendering, large-scale asset curation, built-in data collection pipelines, and a standardized evaluation protocol.
The overall system pipeline is illustrated in Fig.~\ref{fig:pipeline}.

\vspace{-1mm}
\subsection{System Architecture}
\vspace{-1mm}
\label{ssec:fullstack}

\begin{wrapfigure}{r}{0.32\columnwidth}
    \centering
    \includegraphics[width=0.32\columnwidth]{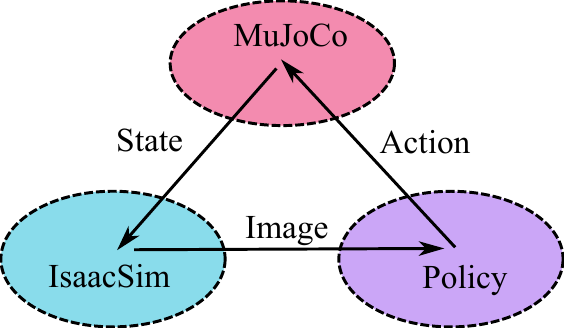}
    \caption{\textbf{System Diagram.} MuJoCo simulates physical interactions, while Isaac Sim synchronizes states and renders photorealistic images for policy inference.}
    \label{fig:system}
    \vspace{-5mm}
\end{wrapfigure}

\ours employs a dual-simulator architecture that strictly decouples physical simulation from visual rendering.
As shown in Fig.~\ref{fig:system}, MuJoCo \cite{todorov2012mujoco} handles all rigid-body dynamics, contact resolution, and robot control.
At each simulation step, the physical states of the robot and environment are synchronized to Isaac Sim \cite{NVIDIA_Isaac_Sim}, which performs photorealistic ray-traced rendering.
This separation allows \ours to harness both the superior contact fidelity and locomotion stability of MuJoCo and the high-quality visual diversity of Isaac Sim.

To support generalist foundation models, \ours implements a decoupled whole-body control scheme.
The high-level policy (\textit{e.g.}, a VLA) predicts kinematic trajectories for the upper body and navigation commands for the base.
These commands are subsequently passed to a state-of-the-art lower-body tracking controller \cite{li2025amo,luo2025sonic,bjorck2025gr00t}, which operates at a higher frequency to maintain balance and execute locomotion.
To ensure seamless integration with existing reinforcement learning workflows, the entire simulation loop is wrapped in a standard OpenAI Gym interface \cite{brockman2016openai}, as detailed in Algorithm \ref{algo:workflow}.

\vspace{-2mm}
\begin{algorithm}
\footnotesize
\caption{Core Policy Evaluation Workflow in \ours}
\begin{algorithmic}[1]
\State import gymnasium as gym \Comment{Gym-style API}
\State env = gym.make(\textit{"}G1WholebodyXmovePickTeleop-v0\textit{"},\, \\
\hspace*{2em} render\_model=\textit{"}mujoco\_isaacsim\textit{"}) \Comment{Create Dual-Sim Env}
\State agent = Psi0Agent(env.robot) \Comment{Instantiate VLA Agent}
\State env = LerobotEnv(env) \Comment{Wrapper for trajectory recording}
\State obs, info = env.reset() \Comment{Initialize environment}
\While{not (terminated or truncated)} \Comment{50 Hz control loop}
    \State action = agent.get\_action(obs, info, instruction)  \Comment{Policy inference}
    \State obs, info, truncated, terminated = env.step(action) \Comment{Advance simulation}
\EndWhile
\end{algorithmic}
\label{algo:workflow}
\end{algorithm}
\vspace{-3mm}

\subsection{Large-Scale Asset Curation and Task Design}
\vspace{-1mm}
\label{ssec:assets}
To ensure sufficient diversity for training robust vision-language-action models, we curated a massive library of objects and environments.
We imported 53 objects from GraspNet-1B \cite{fang2020graspnet} and over 1,500 diverse objects from Objaverse \cite{deitke2023objaverse}.
To guarantee stable physics in MuJoCo, all object meshes underwent convex decomposition via CoACD \cite{wei2022coacd} to generate accurate collision geometries.
Simultaneously, all assets were converted into USD format with high-resolution textures for Isaac Sim rendering.
To move beyond isolated tabletop setups, we integrated 50 complete indoor scenes from the HSSD dataset \cite{khanna2024habitat}.
Using these assets, we designed 60 distinct whole-body loco-manipulation tasks, encompassing rigid object pick-and-place, non-prehensile interactions, and articulated object manipulation.
Detailed illustrations of each task and other entities are included in the class diagrams provided in the Appendix.

\vspace{-1mm}
\subsection{Scalable Data Collection Pipelines}
\vspace{-1mm}
\label{ssec:datagen}
A core feature of \ours is its built-in infrastructure for generating the large-scale demonstration data required to fine-tune humanoid policies.
We provide two distinct pipelines: automated motion planning and human teleoperation.

\begin{figure}[!t]
    \centering
    \includegraphics[width=1\linewidth]{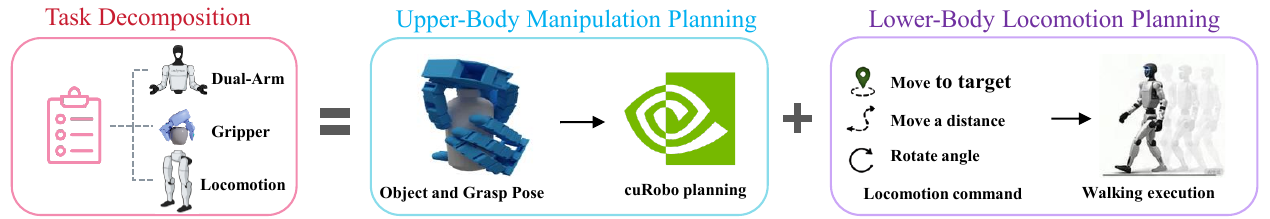}
    \vspace{-6mm}
     \caption{\textbf{Automated Motion Planning.} Based on task decomposition, scripted policies coordinate upper-body manipulation via motion planning and lower-body movement to generate automated demonstrations.}
    \label{fig:mp_pipeline}
    \vspace{-5mm}
\end{figure}

\paragraph{Automated Motion Planning.}
For scalable trajectory generation, we implemented an automated pipeline driven by CuRobo \cite{curobo_v2}, as illustrated in Fig.~\ref{fig:mp_pipeline}.
To ensure physically plausible grasping, objects are first dropped in MuJoCo to determine their stable resting poses.
These stable poses are then processed by BoDex \cite{chen2024bodex} to synthesize feasible grasp configurations.
During task execution, a scripted policy decomposes the high-level objective into atomic actions.
CuRobo generates the kinematic trajectories for the dual arms to execute the grasp, while a scripted lower-body policy coordinates the necessary base movements to reach the target. We detail the grasp synthesis process and trajectory decomposition in the Appendix.

\paragraph{Low-Latency VR Teleoperation.}
Because motion planning struggles with complex dexterous manipulation, we also integrated a human teleoperation pipeline.
To minimize latency, Isaac Sim rendering is disabled during data collection.
Instead, MuJoCo's native renderer streams egocentric stereo video directly to a PICO XR headset.
The operator's hand movements are retargeted to the humanoid's upper body via inverse kinematics, while a whole-body tracking policy \cite{li2025amo, luo2025sonic} autonomously manages the robot's balance and locomotion based on the operator's joystick inputs.
This decoupled approach drastically reduces the operator's cognitive load.
Demonstrations are recorded at 50 Hz and automatically exported into the standardized LeRobot format \cite{cadene2026lerobot}.

\vspace{-1mm}
\subsection{Offline Rendering and Domain Randomization}
\vspace{-1mm}
\label{ssec:render}
To close the visual sim-to-real gap, trajectories collected via motion planning or teleoperation are replayed offline in Isaac Sim.
During this replay phase, we apply extensive domain randomization.
We randomize object instances, initial poses, tabletop textures, lighting conditions, camera viewpoints, and language instructions.
Specifically, we utilize NVIDIA vMaterials \cite{nvidia_vmaterials} to inject high-quality, diverse surface properties into the scenes.
This offline rendering pipeline enables the generation of high-quality RGB observations under diverse visual conditions for downstream policy learning and evaluation.

\vspace{-1mm}
\subsection{Policy Integration and Evaluation Protocol}
\vspace{-1mm}
\label{ssec:eval}
To standardize evaluation, \ours defines a strict protocol with three progressively difficult levels.
Level 1 introduces random distractor objects into the scene.
Level 2 adds visual randomization, altering object materials and environmental lighting to test perceptual robustness.
Level 3 introduces spatial randomization, varying the initial positions of both the target objects and the robot to evaluate layout generalization.

For policy integration, \ours natively supports the $\Psi_0$ \cite{wei2026psi0} training stack.
The standard whole-body representation consists of a 32-dimensional proprioceptive state (14 DoF hands, 14 DoF arms, waist roll-pitch-yaw, and base height).
The action space is 36-dimensional, comprising the 32-dimensional upper-body targets and 4-dimensional locomotion commands ($v_x, v_y, v_{yaw}, q_{yaw}$).
Using this unified interface, we rendered 6,000 episodes to facilitate the large-scale benchmarking of mainstream VLA and WAM architectures.

\vspace{-1mm}
\section{Experiments}
\vspace{-1mm}
In this section, we evaluate the utility and scalability of the \ours benchmark. We first present the results of our large-scale benchmarking, where state-of-the-art VLA models and WAMs are evaluated in a controlled environment across multiple tasks.
We then assess the efficiency of our data generation pipelines.
Next, we conduct ablation studies to analyze the impact of domain randomization, data scaling, and data sources on policy learning.
Finally, we evaluate the feasibility of transferring policies trained entirely in \ours to the real world in a zero-shot manner.

\subsection{Benchmarking of Humanoid Policies}
To benchmark state-of-the-art humanoid models, we select six representative tasks and fine-tune each baseline using 100 demonstrations for each task, and then evaluate them across the three domain-randomization levels defined in Section~\ref{ssec:eval}.

\begin{table}[h]
\centering
\footnotesize
\begin{tabular*}{\textwidth}{@{\extracolsep{\fill}} p{2.2cm} c@{\hspace{0.6em}}c@{\hspace{0.6em}}c@{\hspace{0.6em}}c@{\hspace{0.6em}}c@{\hspace{0.6em}}c @{}}
\thickrule
\tablesp
\textbf{Baseline}
& \textbf{XMovePick}
& \textbf{BendPick}
& \textbf{Handover}
& \textbf{Mobile P\&P}
& \textbf{Grasp}
& \textbf{XMoveBendPick}\\
\tablesp
\thinrule
\tablesp

\textbf{$\Psi_0$} \cite{wei2026psi0}
& 10 / 10 / 6
& \textbf{10 / 10 / 10}
& \textbf{7 / 7 / 10}
& \textbf{7 / 5 / 6}
& \textbf{10 / 10 / 8}
& \ul{10 / 9 / 9} \\

GR00T N1.6 \cite{bjorck2025gr00t}
& \ul{10 / 10 / 7}
& 7 / 7 / 6
& 1 / 3 / 3
& 0 / 0 / 0
& 9 / 9 / 7
& 4 / 4 / 1 \\

$\pi_{0.5}$ \cite{intelligencepi05}
& 7 / 5 / 1
& \ul{10 / 10 / 8}
& 5 / 4 / 5
& 3 / 3 / 3
& \textbf{10 / 10 / 8}
& 0 / 0 / 0 \\

DreamZero \cite{dreamzero}
& \textbf{10 / 10 / 10} 
& 9 / 9 / 8
& \ul{7 / 8 / 9}
& 5 / 3 / 3
& 9 / 10 / 7
& 0 / 0 / 1 \\

Cosmos3 \cite{cosmos3}
& 10 / 10 / 8
& 0 / 0 / 0
& 7 / 10 / 7
& 7 / 5 / 2
& 3 / 3 / 3
& 5 / 2 / 1 \\

DP \cite{dp}
& 3 / 3 / 2
& 10 / 8 / 6
& 3 / 2 / 4
& 4 / 0 / 0
& 8 / 9 / 8
& 0 / 0 / 0 \\

ACT \cite{act}
& 10 / 10 / 5
& 10 / 9 / 9
& \textbf{7 / 7 / 10}
& \ul{5 / 5 / 5}
& \textbf{10 / 10 / 8}
& \textbf{9 / 10 / 10} \\

\tablesp
\thickrule
\end{tabular*}
\caption{
\textbf{Single-Task Fine-Tuning Results.}
Each entry reports performance under three domain-randomization levels
(\textit{Level 0 / Level 1 / Level 2}).
\textbf{Boldface} indicates the best performance, and \ul{underlined} values indicate the second best.
The six task labels are
\texttt{XMovePickTeleop-v0},
\texttt{BendPickMP-v0},
\texttt{HandoverTeleop-v0},
\texttt{LocomotionPickBetweenTablesTeleop-v0},
\texttt{Table\allowbreak topGraspMP-v0}, and
\texttt{XMoveBendPickTeleop-v0},
where the suffix \textit{MP} denotes motion planning and \textit{Teleop} denotes teleoperation.
}
\label{tab:benchmark}
\end{table}

\textbf{Baselines.} The evaluated baseline models are categorized into four groups:
\textbf{(1) VLA Models.} $\Psi_0$ \cite{wei2026psi0} is a 2.5B universal humanoid foundation model specialized for rapid adaptation through fine-tuning.
We also evaluate $\pi_{0.5}$ \cite{intelligencepi05} and GR00T-N1.6 \cite{bjorck2025gr00t}, two seminal VLA models for general robotics, carefully adapting their pre-trained weights to the humanoid embodiment.
\textbf{(2) World-Action Models.}
DreamZero \cite{dreamzero} extends video generation models to simultaneously predict future video frames and actions.
By imagining future interactions during execution, the video prediction mechanism serves as planning in image space.
We also include Cosmos3-Nano \cite{cosmos3}, a recent world-action model that jointly generates video and actions.
\textbf{(3) Imitation Learning Baselines.}
Because our benchmark utilizes single-task fine-tuning, we also include two classic visual imitation learning baselines, Diffusion policy (DP) \cite{dp} and Action Chunk Transformers (ACT) \cite{act}.

\textbf{Experimental Results.} Additional details regarding the adaptation of each baseline are provided in the Appendix. As shown in Table~\ref{tab:benchmark}, $\Psi_0$ consistently performs well across most tasks and leads the benchmark. DreamZero and $\pi_{0.5}$ demonstrate strong generalization performance on \textit{level 2} across most tasks, with the exception of Task 6, likely because it requires highly precise base movements. 
Interestingly, the relatively small ACT model achieves strong overall performance, which we attribute to its data efficiency on \textit{single task overfitting} of the SIMPLE dataset, whose demonstrations are less noisy than real-world teleoperation data.
To complement the single-task finetuning results, multi-task benchmarking results of the foundation models are provided in Appendix~\ref{app:extended}.

\vspace{-1mm}
\subsection{Data Collection Efficiency in the Simulator}
\vspace{-1mm}

A key motivation for building \ours\ is to reduce the cost and effort of collecting high-quality robot demonstration data.
We compare three data collection paradigms across three representative tasks: \textit{whole-body pick-and-place} (T1), \textit{stand-still handover} (T2), and \textit{mobile pick-and-place} (T3).
The three paradigms are:
(1) \textbf{Motion planning (MP)} in simulation, which generates demonstrations autonomously without a human operator;
(2) \textbf{Real-robot teleoperation}, where an operator collects demonstrations directly on physical hardware;
and (3) \textbf{Sim teleoperation}, where a human operator controls the robot inside \ours\ via the VR interface;

\begin{table}[!t]
\centering
\footnotesize
\resizebox{\textwidth}{!}{%
\begin{tabular}{lcccccc}
\thickrule
\tablesp
& \multicolumn{2}{c}{\textbf{T1: Whole-Body Pick-Place}}
& \multicolumn{2}{c}{\textbf{T2: Stand-Still Handover}}
& \multicolumn{2}{c}{\textbf{T3: Mobile Pick-Place}} \\
\tablesp
\textbf{Collection Method}
  & \textbf{Demos/hr} $\uparrow$ & \textbf{Avg.\ time (s)} $\downarrow$
  & \textbf{Demos/hr} $\uparrow$ & \textbf{Avg.\ time (s)} $\downarrow$
  & \textbf{Demos/hr} $\uparrow$ & \textbf{Avg.\ time (s)} $\downarrow$ \\
\tablesp
\thinrule
\tablesp
Motion Planning (Sim) & 58.9   & 61.1  & 32.7 & 109.8 & 24.0  & 150.0   \\
Teleoperation (Real)  & \ul{206.8}  & \ul{17.4} & \ul{130.9} & \url{27.5} & \ul{87.2}  & \ul{41.3} \\
Teleoperation (Sim)   & \textbf{310.3}  & \textbf{11.6} & \textbf{197.8} & \textbf{18.2} & \textbf{156.5} & \textbf{23.0} \\
\tablesp
\thickrule
\end{tabular}%
}
\caption{\textbf{Data Collection Efficiency.} Each cell reports the number of demonstrations collected per hour and the mean episode duration in seconds. Motion planning runs without an operator; teleoperation figures reflect a single experienced operator per session.}
\label{tab:data_efficiency}
\end{table}

We measure collection throughput in demonstrations per hour (demos/hr) and average episode duration (seconds), as reported in Table~\ref{tab:data_efficiency}.
Motion planning achieves the lowest throughput due to the difficulty of optimization in high-dimensional spaces; however, it does not incur operator fatigue or hardware reset overhead.
Real-robot teleoperation achieves higher efficiency and captures more natural whole-body motions than motion-planning-based scripted policies.
Conversely, Sim teleoperation is substantially faster because it avoids physical reset overhead, safety constraints, and hardware maintenance.
Beyond raw time efficiency, simulated data collection is fundamentally more scalable.
It eliminates the need for physical humanoid hardware, requiring only a standard VR headset, and allows collected trajectories to be infinitely scaled across diverse environments, lighting conditions, and object instances via offline replay and rendering.

\vspace{-1mm}
\subsection{Ablation Studies}
\vspace{-1mm}
\paragraph{Domain Randomization and Data Scaling.}
Domain randomization (DR) is central to bridging the visual sim-to-real gap.
We ablate the effect of training data composition on cross-level generalization using $\Psi_0$ fine-tuned for 2{,}000 steps.
As reported in Table~\ref{tab:abl_dr_and_scale}, training on Level~0 data alone yields a strong in-distribution success rate (0.80 at Eval Set~0) but generalizes poorly to harder visual conditions (0.50 at Eval Set~2).
Mixing Level~0 and Level~1 training data maintains in-distribution performance while substantially improving out-of-distribution robustness (0.80 on Set~2).
This confirms that exposure to diverse visual conditions during training is necessary for policies to transfer to challenging environments.
The results motivate the mixed-DR evaluation strategy adopted in the main benchmark (Table~\ref{tab:benchmark}). 
Furthermore, we found that increasing the amount of teleoperation data significantly improves performance across all three validation sets.

\begin{table}[!t]
\centering
\footnotesize
\renewcommand{\arraystretch}{1.12}
\setlength{\tabcolsep}{3pt}
\begin{tabular*}{\textwidth}{@{\extracolsep{\fill}} l p{3.2cm} c c c @{}}
\thickrule
\tablesp
\textbf{Task}
& \textbf{Training data}
& \textbf{Eval Set 0}
& \textbf{Eval Set 1}
& \textbf{Eval Set 2} \\
\tablesp
\thinrule
\tablesp
\multirow{2}{*}{BendHandover}
& 10$\times$ Level 0
    & \textbf{8/10 = 0.80}
    & \textbf{8/10 = 0.80}
    & 5/10 = 0.50 \\

& 5$\times$ Level 0 + 5$\times$ Level 1
    & \textbf{8/10 = 0.80}
    & 7/10 = 0.70
    & \textbf{8/10 = 0.80} \\
\midrule
\multirow{2}{*}{XmoveBendPick}
& 10 Trajectories (Teleop)
    & 5/10 = 0.50
    & 6/10 = 0.60
    & 3/10 = 0.30 \\

& 100 Trajectories (Teleop)
    & \textbf{10/10 = 1.00}
    & \textbf{9/10 = 0.90}
    & \textbf{9/10 = 0.90} \\
\tablesp
\thickrule
\end{tabular*}
\caption{
\textbf{Ablation Study on Domain Randomization.}
Evaluation success rates of $\Psi_{0}$ fine-tuned for 2,000 training steps under different training data compositions.
Training with mixed domain-randomization levels improves generalization to harder evaluation settings.
Scaling up teleoperation data further improves performance.
}
\label{tab:abl_dr_and_scale}
\vspace{-5mm}
\end{table}

\paragraph{Data Source Quality.} 
To study the impact of different data sources, we fine-tune $\Psi_0$ \cite{wei2026psi0} on three tasks using either motion-planning-only data or teleoperation-only data (Table \ref{tab:abl_data}).
In general, human teleoperation data proves more suitable for learning complex tasks, which we attribute to its diverse and naturalistic motion profiles.

\begin{table}[!t]
\centering
\footnotesize
\setlength{\tabcolsep}{2pt}
\begin{minipage}{0.53\textwidth}
\centering
\begin{tabular*}{\textwidth}{@{\extracolsep{\fill}} l c c c c @{}}
\thickrule
\tablesp
\textbf{Variant} & \textbf{BendPick} & \textbf{Mobile P\&P} & \textbf{XMoveBendPick} & \textbf{Avg.} \\
\tablesp
\thinrule
\tablesp

MP
    & \textbf{10/10/10}
    & 3/2/2
    & 4/2/2
    & 5.00 \\

Teleop
    & 8/8/6
    & \textbf{7/5/6}
    & \textbf{10/9/9}
    & \textbf{7.56} \\

\tablesp
\thickrule
\end{tabular*}
\caption{
\textbf{Ablation Study on Data Source.}
We compare motion-planning-only and teleoperation-only training data across three task families. Teleoperation data leads to better performance. 
}
\label{tab:abl_data}
\end{minipage}\hfill
\begin{minipage}{0.45\textwidth}
\centering
\begin{tabular*}{\textwidth}{@{\extracolsep{\fill}} l c c @{}}
\thickrule
\tablesp
\textbf{Task}
& \textbf{Sim Eval}
& \textbf{Real Eval} \\
\tablesp
\thinrule
\tablesp
Pick \& Place
  & 9/10 = 0.90
  & 8/10 = 0.80 \\
Handover
  & 10/10 = 1.00
  & 8/10 = 0.80 \\
XMoveBendCarryBox
  & 7/10 = 0.70
  & 5/10 = 0.50 \\
\tablesp
\thickrule
\end{tabular*}
\caption{
\textbf{Zero-Shot Sim-to-Real Transfer.}
Fine-tuned exclusively on simulation data, evaluated both in the simulator and directly in the real world.
}
\label{tab:sim2real-success}
\end{minipage}
    \vspace{-7mm}
\end{table}

\subsection{Zero-Shot Sim-to-Real Transfer}
\begin{figure}[htbp]
    \centering
    \includegraphics[width=0.88\textwidth]{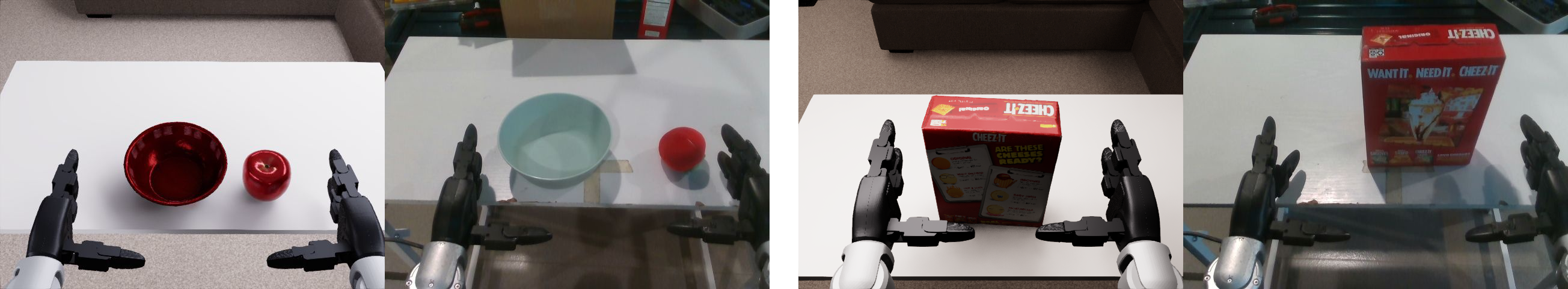}
    \vspace{-2mm}
    \caption{
    \textbf{Sim-to-Real Task Observations.}
    Each task is shown as an adjacent simulation--real pair, with the simulated egocentric view on the left and the corresponding real egocentric view on the right.
    The left pair shows the pick-and-place task, and the right pair shows the handover task.
    }
    \vspace{-6mm}
    \label{fig:sim2real-tasks}
\end{figure}

While \ours is primarily designed as a simulation benchmark, we investigate whether policies trained exclusively on \ours data can transfer to the real world.
Because real-world evaluation is highly constrained, we perform this evaluation on a subset of tasks under similar environmental settings, as illustrated in Figure~\ref{fig:sim2real-tasks}.
As shown in Table~\ref{tab:sim2real-success}, we observe that the learned $\Psi_0$ policy generalizes to the real world in a zero-shot fashion without requiring real-world data for downstream fine-tuning.
This suggests that the high-fidelity physics and photorealistic rendering provided by \ours are sufficient to cross the sim-to-real gap for certain loco-manipulation tasks.

\vspace{-1mm}
\subsection{Whole-Body Task Coverage}
\label{ssec:wbc_tasks}
\begin{figure}[htbp]
    \centering
    \includegraphics[width=0.68\textwidth]{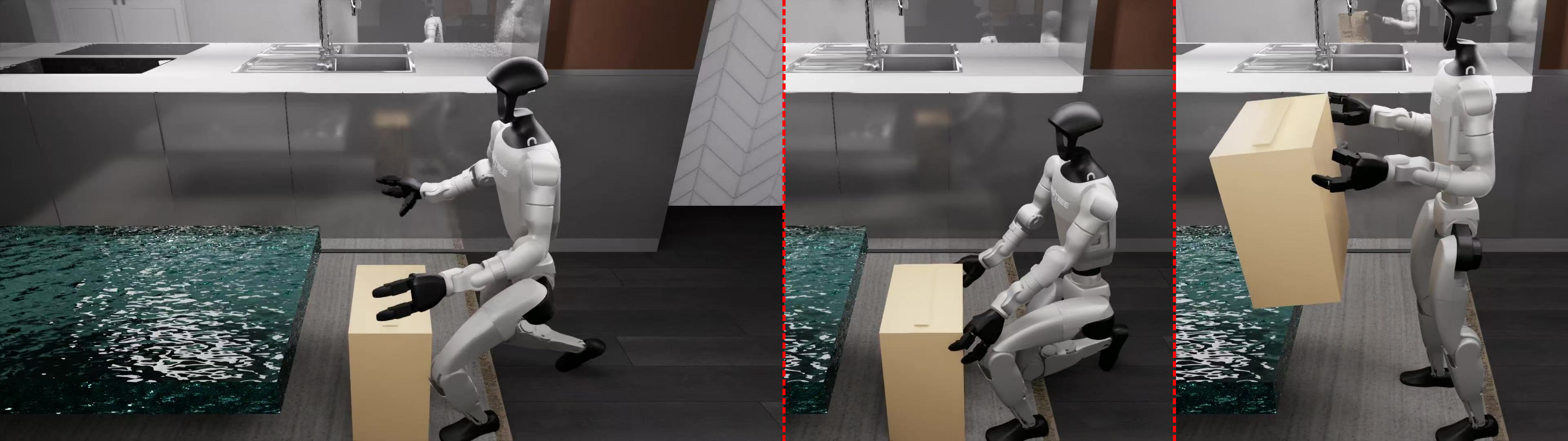}
    \vspace{-3mm}
    \caption{
    \textbf{Whole-Body Task Example.}
    Three frames of \textit{XMoveBendCarryBox}: the robot approaches the box, kneels to wrap both arms around it, and lifts it onto the table.
    }
    \vspace{-4mm}
    \label{fig:wbc}
\end{figure}

\vspace{-1mm}
Beyond tabletop manipulation, \ours includes whole-body tasks that require coordinated arm--leg motion, such as \textit{XMovePushBox}, \textit{KickBox}, and \textit{XMoveBendCarryBox}.
Figure~\ref{fig:wbc} illustrates \textit{XMoveBendCarryBox}, where the robot kneels to lower its torso and then wraps both arms around the box to lift it onto the table.
The same task is also included in our sim-to-real evaluation (Table~\ref{tab:sim2real-success}).

\vspace{-1mm}
\subsection{Physics--Rendering Consistency}
\label{ssec:consistency}
\begin{figure}[htbp]
    \centering
    \includegraphics[width=0.68\textwidth]{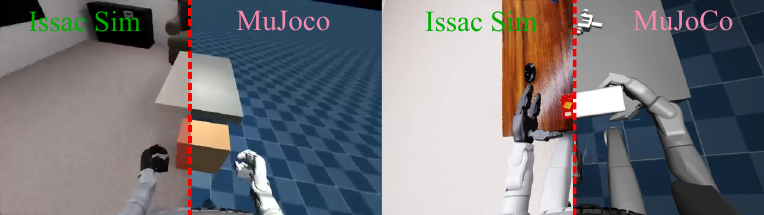}
    \vspace{-3mm}
    \caption{
    \textbf{Visual--Physics Mesh Alignment.}
    Two egocentric frames, each shown as a pair with the Isaac Sim visual rendering on the left and the corresponding MuJoCo physics rendering on the right.
    }
    \vspace{-4mm}
    \label{fig:align}
\end{figure}

\vspace{-1mm}
Because physics and rendering run in separate engines, we verify that the collision meshes simulated in MuJoCo stay aligned with the visual meshes rendered in Isaac Sim.
Figure~\ref{fig:align} compares the two from the robot's egocentric view; they are closely aligned across frames, as evidenced by the consistent table surfaces, cuboid object edges, and humanoid torso.
We also note that the $\sim$4\,fps rendering throughput is bound by the \textit{synchronized} render mode of Isaac Sim.
It can be mitigated by parallelizing environments during offline data generation or policy inference, in which case throughput is bottlenecked only by available compute.

\vspace{-1mm}
\subsection{Simulation Parameters in MuJoCo}
\label{ssec:sim_params}
\vspace{-1mm}
Table~\ref{tab:sim_params} lists the main tuned simulation parameters, covering actuator, solver, and contact settings, except for the control gains, which vary with the whole-body controller (AMO or SONIC).
These were extensively tested to ensure stable and realistic simulation:
(1) maximum force limits are enforced for all leg, arm, and finger joints;
(2) joint friction introduces a dead zone that prevents small torques from inducing unintended motion;
(3) viscous damping suppresses joint oscillations;
and (4) joint armature adds rotational inertia to improve numerical stability.
\begin{table}[htbp]
\centering
\footnotesize
\setlength{\tabcolsep}{4pt}
\begin{tabular*}{\textwidth}{@{\extracolsep{\fill}} l l l l l l @{}}
\thickrule
\tablesp
\textbf{Parameter} & \textbf{Default} & \textbf{Type} & \textbf{Parameter} & \textbf{Default} & \textbf{Type} \\
\tablesp
\thinrule
\tablesp
\texttt{actuatorfrclimited} & true         & Actuator & \texttt{impratio} & 10             & Solver  \\
\texttt{frictionloss}       & 0.2          & Actuator & \texttt{cone}     & elliptic       & Contact \\
\texttt{damping}            & 2            & Actuator & \texttt{solref}   & 0.005 1        & Contact \\
\texttt{armature}           & 0.05         & Actuator & \texttt{solimp}   & 0.9 0.95 0.001 & Contact \\
\texttt{integrator}         & implicitfast & Solver   & $K_p$/$K_d$       & \textit{varied} & Control \\
\tablesp
\thickrule
\end{tabular*}
\caption{
\textbf{Simulation Parameters in MuJoCo.}
Main tuned physics and actuator parameters shared across all tasks. Control gains vary with the whole-body controller (AMO or SONIC).
}
\label{tab:sim_params}
\vspace{-7mm}
\end{table}

\vspace{-2mm}
\section{Conclusion}
\vspace{-1mm}
\label{sec:conclusion}
We presented \ours, a full-stack simulation infrastructure designed to standardize the evaluation and training of humanoid policies.
By coupling the robust contact physics of MuJoCo with the photorealistic rendering of Isaac Sim, \ours provides a high-fidelity environment tailored for whole-body loco-manipulation.
The framework features a large-scale benchmark comprising \ntask diverse tasks, over 1,000 objects, and 50 indoor scenes.
To facilitate scalable policy learning, \ours integrates built-in data collection pipelines for both automated motion planning and low-latency VR teleoperation.
Our extensive benchmarking of state-of-the-art VLA and WAM architectures demonstrates that performance in \ours strongly correlates with real-world outcomes, and that data generated in SIMPLE empowers effective sim-to-real transfer.
We hope \ours will serve as a reproducible and scalable foundation to accelerate future research in humanoid robotics.

\paragraph{Limitations.}
Despite these contributions, \ours\ has several limitations that point toward future work. 
\textit{(1) Rendering throughput.}
The photorealistic ray-tracing pipeline in Isaac Sim is computationally expensive.
A single GPU renders $\sim$4 frames per second, making large-scale dataset generation time-intensive.
\textit{(2) Rigid-body assumption.}
The current simulation pipeline models all objects as rigid bodies.
Deformable and soft-body objects such as cloth, rope, or food items cannot be faithfully simulated within the current framework.

\clearpage

\acknowledgments{
We thank Weiduo Yuan for assistance in setting up the multi-task fine-tuning of DreamZero, Jingxuan Wang and Junsoo Kim for support in integrating the SONIC whole-body controller, and Andrew Park for their help in testing \ours.
}

\bibliography{ref}

\begin{thebibliography}{44}
\providecommand{\natexlab}[1]{#1}
\providecommand{\url}[1]{\texttt{#1}}
\expandafter\ifx\csname urlstyle\endcsname\relax
  \providecommand{\doi}[1]{doi: #1}\else
  \providecommand{\doi}{doi: \begingroup \urlstyle{rm}\Url}\fi

\bibitem[Bjorck et~al.(2025)Bjorck, Casta{\~n}eda, Cherniadev, Da, Ding, Fan,
  Fang, Fox, Hu, Huang, et~al.]{bjorck2025gr00t}
J.~Bjorck, F.~Casta{\~n}eda, N.~Cherniadev, X.~Da, R.~Ding, L.~Fan, Y.~Fang,
  D.~Fox, F.~Hu, S.~Huang, et~al.
\newblock Gr00t n1: An open foundation model for generalist humanoid robots.
\newblock \emph{arXiv preprint arXiv:2503.14734}, 2025.

\bibitem[Ye et~al.(2026)Ye, Ge, Zheng, Gao, Yu, Kurian, Indupuru, Tan, Zhu,
  Xiang, et~al.]{dreamzero}
S.~Ye, Y.~Ge, K.~Zheng, S.~Gao, S.~Yu, G.~Kurian, S.~Indupuru, Y.~L. Tan,
  C.~Zhu, J.~Xiang, et~al.
\newblock World action models are zero-shot policies.
\newblock \emph{arXiv preprint arXiv:2602.15922}, 2026.

\bibitem[Intelligence et~al.(2025)Intelligence, Black, Brown, Darpinian,
  Dhabalia, Driess, Esmail, Equi, Finn, Fusai, et~al.]{intelligencepi05}
P.~Intelligence, K.~Black, N.~Brown, J.~Darpinian, K.~Dhabalia, D.~Driess,
  A.~Esmail, M.~Equi, C.~Finn, N.~Fusai, et~al.
\newblock {$\pi_{0.5}$}: A vision-language-action model with open-world
  generalization.
\newblock \emph{arXiv preprint arXiv:2504.16054}, 2025.

\bibitem[Wei et~al.(2026)Wei, Jing, Li, Zhao, Mao, Ni, He, Liu, Liu, Kang,
  et~al.]{wei2026psi0}
S.~Wei, H.~Jing, B.~Li, Z.~Zhao, J.~Mao, Z.~Ni, S.~He, J.~Liu, X.~Liu, K.~Kang,
  et~al.
\newblock {$\Psi_0$}: An open foundation model towards universal humanoid
  loco-manipulation.
\newblock \emph{arXiv preprint arXiv:2603.12263}, 2026.

\bibitem[Chen et~al.(2025)Chen, Chen, Chen, Cai, Liu, Li, Liang, Lin, Ge, Gu,
  et~al.]{chen2025robotwin}
T.~Chen, Z.~Chen, B.~Chen, Z.~Cai, Y.~Liu, Z.~Li, Q.~Liang, X.~Lin, Y.~Ge,
  Z.~Gu, et~al.
\newblock Robotwin 2.0: A scalable data generator and benchmark with strong
  domain randomization for robust bimanual robotic manipulation.
\newblock \emph{arXiv preprint arXiv:2506.18088}, 2025.

\bibitem[Geng et~al.(2025)Geng, Wang, Wei, Li, Wang, An, Cheng, Lou, Li, Wang,
  et~al.]{geng2025roboverse}
H.~Geng, F.~Wang, S.~Wei, Y.~Li, B.~Wang, B.~An, C.~T. Cheng, H.~Lou, P.~Li,
  Y.-J. Wang, et~al.
\newblock Roboverse: Towards a unified platform, dataset and benchmark for
  scalable and generalizable robot learning.
\newblock \emph{arXiv preprint arXiv:2504.18904}, 2025.

\bibitem[Liu et~al.(2023)Liu, Zhu, Gao, Feng, Liu, Zhu, and
  Stone]{liu2023libero}
B.~Liu, Y.~Zhu, C.~Gao, Y.~Feng, Q.~Liu, Y.~Zhu, and P.~Stone.
\newblock Libero: Benchmarking knowledge transfer for lifelong robot learning.
\newblock \emph{Advances in Neural Information Processing Systems},
  36:\penalty0 44776--44791, 2023.

\bibitem[Li et~al.(2023)Li, Zhang, Wong, Gokmen, Srivastava,
  Mart{\'\i}n-Mart{\'\i}n, Wang, Levine, Lingelbach, Sun,
  et~al.]{li2023behavior}
C.~Li, R.~Zhang, J.~Wong, C.~Gokmen, S.~Srivastava, R.~Mart{\'\i}n-Mart{\'\i}n,
  C.~Wang, G.~Levine, M.~Lingelbach, J.~Sun, et~al.
\newblock Behavior-1k: A benchmark for embodied ai with 1,000 everyday
  activities and realistic simulation.
\newblock In \emph{Conference on Robot Learning}, pages 80--93. PMLR, 2023.

\bibitem[Atreya et~al.(2025)Atreya, Pertsch, Lee, Kim, Jain, Kuramshin, Eppner,
  Neary, Hu, Ramos, et~al.]{atreya2025roboarena}
P.~Atreya, K.~Pertsch, T.~Lee, M.~J. Kim, A.~Jain, A.~Kuramshin, C.~Eppner,
  C.~Neary, E.~Hu, F.~Ramos, et~al.
\newblock Roboarena: Distributed real-world evaluation of generalist robot
  policies.
\newblock In \emph{Proceedings of the Conference on Robot Learning (CoRL
  2025)}, 2025.

\bibitem[Zhou et~al.(2025)Zhou, Atreya, Tan, Pertsch, and
  Levine]{zhou2025autoeval}
Z.~Zhou, P.~Atreya, Y.~L. Tan, K.~Pertsch, and S.~Levine.
\newblock Autoeval: Autonomous evaluation of generalist robot manipulation
  policies in the real world.
\newblock \emph{arXiv preprint arXiv:2503.24278}, 2025.

\bibitem[Li et~al.(2024)Li, Hsu, Gu, Pertsch, Mees, Walke, Fu, Lunawat, Sieh,
  Kirmani, et~al.]{li2024simplerenv}
X.~Li, K.~Hsu, J.~Gu, K.~Pertsch, O.~Mees, H.~R. Walke, C.~Fu, I.~Lunawat,
  I.~Sieh, S.~Kirmani, et~al.
\newblock Evaluating real-world robot manipulation policies in simulation.
\newblock \emph{arXiv preprint arXiv:2405.05941}, 2024.

\bibitem[Lab(2025)]{humanoidverse}
C.~L. Lab.
\newblock Humanoidverse: A multi-simulator framework for humanoid robot
  sim-to-real learning.
\newblock \url{https://github.com/LeCAR-Lab/HumanoidVerse}, 2025.

\bibitem[Sferrazza et~al.(2024)Sferrazza, Huang, Lin, Lee, and
  Abbeel]{sferrazza2024humanoidbench}
C.~Sferrazza, D.-M. Huang, X.~Lin, Y.~Lee, and P.~Abbeel.
\newblock Humanoidbench: Simulated humanoid benchmark for whole-body locomotion
  and manipulation, 2024.

\bibitem[Khanna et~al.(2024)Khanna, Mao, Jiang, Haresh, Shacklett, Batra,
  Clegg, Undersander, Chang, and Savva]{khanna2024habitat}
M.~Khanna, Y.~Mao, H.~Jiang, S.~Haresh, B.~Shacklett, D.~Batra, A.~Clegg,
  E.~Undersander, A.~X. Chang, and M.~Savva.
\newblock Habitat synthetic scenes dataset (hssd-200): An analysis of 3d scene
  scale and realism tradeoffs for objectgoal navigation.
\newblock In \emph{Proceedings of the IEEE/CVF Conference on Computer Vision
  and Pattern Recognition}, pages 16384--16393, 2024.

\bibitem[Deitke et~al.(2023)Deitke, Schwenk, Salvador, Weihs, Michel,
  VanderBilt, Schmidt, Ehsani, Kembhavi, and Farhadi]{deitke2023objaverse}
M.~Deitke, D.~Schwenk, J.~Salvador, L.~Weihs, O.~Michel, E.~VanderBilt,
  L.~Schmidt, K.~Ehsani, A.~Kembhavi, and A.~Farhadi.
\newblock Objaverse: A universe of annotated 3d objects.
\newblock In \emph{Proceedings of the IEEE/CVF conference on computer vision
  and pattern recognition}, pages 13142--13153, 2023.

\bibitem[Todorov et~al.(2012)Todorov, Erez, and Tassa]{todorov2012mujoco}
E.~Todorov, T.~Erez, and Y.~Tassa.
\newblock Mujoco: A physics engine for model-based control.
\newblock In \emph{2012 IEEE/RSJ international conference on intelligent robots
  and systems}, pages 5026--5033. IEEE, 2012.

\bibitem[{NVIDIA}()]{NVIDIA_Isaac_Sim}
{NVIDIA}.
\newblock {Isaac Sim}.
\newblock URL \url{https://github.com/isaac-sim/IsaacSim}.

\bibitem[Sundaralingam et~al.(2026)Sundaralingam, Murali, and
  Birchfield]{curobo_v2}
B.~Sundaralingam, A.~Murali, and S.~Birchfield.
\newblock curobov2: Dynamics-aware motion generation with depth-fused distance
  fields for high-dof robots, 2026.

\bibitem[Chen et~al.(2024)Chen, Ke, and Wang]{chen2024bodex}
J.~Chen, Y.~Ke, and H.~Wang.
\newblock Bodex: Scalable and efficient robotic dexterous grasp synthesis using
  bilevel optimization.
\newblock \emph{arXiv preprint arXiv:2412.16490}, 2024.

\bibitem[Li et~al.(2025)Li, Cheng, Huang, Yang, Qiu, and Wang]{li2025amo}
J.~Li, X.~Cheng, T.~Huang, S.~Yang, R.~Qiu, and X.~Wang.
\newblock Amo: Adaptive motion optimization for hyper-dexterous humanoid
  whole-body control.
\newblock \emph{Robotics: Science and Systems 2025}, 2025.

\bibitem[Luo et~al.(2025)Luo, Yuan, Wang, Li, Chen, Castaneda, Cao, Li, Minor,
  Ben, et~al.]{luo2025sonic}
Z.~Luo, Y.~Yuan, T.~Wang, C.~Li, S.~Chen, F.~Castaneda, Z.-A. Cao, J.~Li,
  D.~Minor, Q.~Ben, et~al.
\newblock Sonic: Supersizing motion tracking for natural humanoid whole-body
  control.
\newblock \emph{arXiv preprint arXiv:2511.07820}, 2025.

\bibitem[Yang et~al.(2025)Yang, Yu, Wu, Yan, Li, Cheng, Zou, Fang, Cheng, Qiu,
  et~al.]{yang2025egovla}
R.~Yang, Q.~Yu, Y.~Wu, R.~Yan, B.~Li, A.-C. Cheng, X.~Zou, Y.~Fang, X.~Cheng,
  R.-Z. Qiu, et~al.
\newblock Egovla: Learning vision-language-action models from egocentric human
  videos.
\newblock \emph{arXiv preprint arXiv:2507.12440}, 2025.

\bibitem[Bi et~al.(2026)Bi, Wu, Lin, Tan, Su, Su, and Zhu]{bi2026h}
H.~Bi, L.~Wu, T.~Lin, H.~Tan, Z.~Su, H.~Su, and J.~Zhu.
\newblock H-rdt: Human manipulation enhanced bimanual robotic manipulation.
\newblock In \emph{Proceedings of the AAAI Conference on Artificial
  Intelligence}, volume~40, pages 18135--18143, 2026.

\bibitem[Deng et~al.(2025)Deng, Yan, Wei, Ma, Yang, Chen, Zhang, Yang, Zhang,
  Zhang, et~al.]{deng2025graspvla}
S.~Deng, M.~Yan, S.~Wei, H.~Ma, Y.~Yang, J.~Chen, Z.~Zhang, T.~Yang, X.~Zhang,
  W.~Zhang, et~al.
\newblock Graspvla: a grasping foundation model pre-trained on billion-scale
  synthetic action data.
\newblock \emph{arXiv preprint arXiv:2505.03233}, 2025.

\bibitem[Zhang et~al.(2024)Zhang, Wang, Wang, Li, Liu, Wei, Wang, Zhang, and
  Wang]{zhang2024uni}
J.~Zhang, K.~Wang, S.~Wang, M.~Li, H.~Liu, S.~Wei, Z.~Wang, Z.~Zhang, and
  H.~Wang.
\newblock Uni-navid: A video-based vision-language-action model for unifying
  embodied navigation tasks.
\newblock \emph{arXiv preprint arXiv:2412.06224}, 2024.

\bibitem[Fang et~al.(2020)Fang, Wang, Gou, and Lu]{fang2020graspnet}
H.-S. Fang, C.~Wang, M.~Gou, and C.~Lu.
\newblock Graspnet-1billion: A large-scale benchmark for general object
  grasping.
\newblock In \emph{Proceedings of the IEEE/CVF conference on computer vision
  and pattern recognition}, pages 11444--11453, 2020.

\bibitem[Wei et~al.(2024)Wei, Geng, Chen, Deng, Wenbo, Zhao, Fang, Guibas, and
  Wang]{wei2024d}
S.~Wei, H.~Geng, J.~Chen, C.~Deng, C.~Wenbo, C.~Zhao, X.~Fang, L.~Guibas, and
  H.~Wang.
\newblock D3roma: Disparity diffusion-based depth sensing for material-agnostic
  robotic manipulation.
\newblock In \emph{8th Annual Conference on Robot Learning}, 2024.
\newblock URL \url{https://openreview.net/forum?id=7E3JAys1xO}.

\bibitem[Yakefu et~al.(2025)Yakefu, Xie, Xu, Zhang, Zhou, Jia, Yang, Fan,
  Zhang, Peng, et~al.]{yakefu2025robochallenge}
A.~Yakefu, B.~Xie, C.~Xu, E.~Zhang, E.~Zhou, F.~Jia, H.~Yang, H.~Fan, H.~Zhang,
  H.~Peng, et~al.
\newblock Robochallenge: Large-scale real-robot evaluation of embodied
  policies.
\newblock \emph{arXiv preprint arXiv:2510.17950}, 2025.

\bibitem[Khazatsky et~al.(2024)Khazatsky, Pertsch, Nair, Balakrishna, and
  et~al.]{khazatsky2024droid}
A.~Khazatsky, K.~Pertsch, S.~Nair, A.~Balakrishna, and S.~D. et~al.
\newblock {DROID: A Large-Scale In-The-Wild Robot Manipulation Dataset}.
\newblock In \emph{Proceedings of Robotics: Science and Systems (RSS)}, 2024.
\newblock \doi{10.15607/RSS.2024.XX.120}.

\bibitem[Wang et~al.(2025)Wang, Ung, Tannert, Duan, Li, Le, Oswal, Grotz,
  Pumacay, Deng, et~al.]{wang2025roboeval}
Y.~R. Wang, C.~Ung, G.~Tannert, J.~Duan, J.~Li, A.~Le, R.~Oswal, M.~Grotz,
  W.~Pumacay, Y.~Deng, et~al.
\newblock Roboeval: Where robotic manipulation meets structured and scalable
  evaluation.
\newblock \emph{arXiv preprint arXiv:2507.00435}, 2025.

\bibitem[James et~al.(2020)James, Ma, Arrojo, and Davison]{james2020rlbench}
S.~James, Z.~Ma, D.~R. Arrojo, and A.~J. Davison.
\newblock Rlbench: The robot learning benchmark \& learning environment.
\newblock \emph{IEEE Robotics and Automation Letters}, 5\penalty0 (2):\penalty0
  3019--3026, 2020.

\bibitem[Nasiriany et~al.(2026)Nasiriany, Nasiriany, Maddukuri, and
  Zhu]{robocasa365}
S.~Nasiriany, S.~Nasiriany, A.~Maddukuri, and Y.~Zhu.
\newblock Robocasa365: A large-scale simulation framework for training and
  benchmarking generalist robots.
\newblock In \emph{International Conference on Learning Representations
  (ICLR)}, 2026.

\bibitem[Kim et~al.(2026)Kim, Pumacay, Rayyan, Argus, Han, VanderBilt,
  Salvador, Deshpande, Hendrix, Jauhri, Liu, Shafiullah, Guru, Guru, Eftekhar,
  Farley, Clay, Duan, Wolters, Herrasti, Lee, Chalvatzaki, Cui, Farhadi, Fox,
  and Krishna]{molmospaces2026}
Y.~Kim, W.~Pumacay, O.~Rayyan, M.~Argus, W.~Han, E.~VanderBilt, J.~Salvador,
  A.~Deshpande, R.~Hendrix, S.~Jauhri, S.~Liu, N.~M.~M. Shafiullah, M.~Guru,
  A.~Guru, A.~Eftekhar, K.~Farley, D.~Clay, J.~Duan, P.~Wolters, A.~Herrasti,
  Y.-C. Lee, G.~Chalvatzaki, Y.~Cui, A.~Farhadi, D.~Fox, and R.~Krishna.
\newblock Molmospaces: A large-scale open ecosystem for robot navigation and
  manipulation, 2026.

\bibitem[Shukla et~al.(2024)Shukla, Tao, and Su]{shukla2024maniskill}
A.~Shukla, S.~Tao, and H.~Su.
\newblock Maniskill-hab: A benchmark for low-level manipulation in home
  rearrangement tasks.
\newblock \emph{arXiv preprint arXiv:2412.13211}, 2024.

\bibitem[Mittal et~al.(2025)Mittal, Roth, Tigue, Richard, Zhang, Du,
  Serrano-Munoz, Yao, Zurbr{\"u}gg, Rudin, et~al.]{mittal2025isaaclab}
M.~Mittal, P.~Roth, J.~Tigue, A.~Richard, O.~Zhang, P.~Du, A.~Serrano-Munoz,
  X.~Yao, R.~Zurbr{\"u}gg, N.~Rudin, et~al.
\newblock Isaac lab: A gpu-accelerated simulation framework for multi-modal
  robot learning.
\newblock \emph{arXiv preprint arXiv:2511.04831}, 2025.

\bibitem[Chen et~al.(2025)Chen, Wei, Xiao, Lyu, Chen, Zhu, and
  Wang]{chen2025robohanger}
Y.~Chen, S.~Wei, B.~Xiao, J.~Lyu, J.~Chen, F.~Zhu, and H.~Wang.
\newblock Robohanger: Learning generalizable robotic hanger insertion for
  diverse garments.
\newblock \emph{IEEE Robotics and Automation Letters}, 2025.

\bibitem[Makoviychuk et~al.(2021)Makoviychuk, Wawrzyniak, Guo, Lu, Storey,
  Macklin, Hoeller, Rudin, Allshire, Handa, et~al.]{makoviychuk2021isaacgym}
V.~Makoviychuk, L.~Wawrzyniak, Y.~Guo, M.~Lu, K.~Storey, M.~Macklin,
  D.~Hoeller, N.~Rudin, A.~Allshire, A.~Handa, et~al.
\newblock Isaac gym: High performance gpu-based physics simulation for robot
  learning.
\newblock \emph{arXiv preprint arXiv:2108.10470}, 2021.

\bibitem[Brockman et~al.(2016)Brockman, Cheung, Pettersson, Schneider,
  Schulman, Tang, and Zaremba]{brockman2016openai}
G.~Brockman, V.~Cheung, L.~Pettersson, J.~Schneider, J.~Schulman, J.~Tang, and
  W.~Zaremba.
\newblock Openai gym.
\newblock \emph{arXiv preprint arXiv:1606.01540}, 2016.

\bibitem[Wei et~al.(2022)Wei, Liu, Ling, and Su]{wei2022coacd}
X.~Wei, M.~Liu, Z.~Ling, and H.~Su.
\newblock Approximate convex decomposition for 3d meshes with collision-aware
  concavity and tree search.
\newblock \emph{ACM Transactions on Graphics (TOG)}, 41\penalty0 (4):\penalty0
  1--18, 2022.

\bibitem[Cadene et~al.(2026)Cadene, Aliberts, Capuano, Aractingi, Zouitine,
  Kooijmans, Choghari, Russi, Pascal, Palma, et~al.]{cadene2026lerobot}
R.~Cadene, S.~Aliberts, F.~Capuano, M.~Aractingi, A.~Zouitine, P.~Kooijmans,
  J.~Choghari, M.~Russi, C.~Pascal, S.~Palma, et~al.
\newblock Lerobot: An open-source library for end-to-end robot learning.
\newblock \emph{arXiv preprint arXiv:2602.22818}, 2026.

\bibitem[{NVIDIA}(2026{\natexlab{a}})]{nvidia_vmaterials}
{NVIDIA}.
\newblock vmaterials.
\newblock \url{https://developer.nvidia.com/vmaterials}, 2026{\natexlab{a}}.
\newblock Accessed: 2026-05-24.

\bibitem[{NVIDIA}(2026{\natexlab{b}})]{cosmos3}
{NVIDIA}.
\newblock Cosmos 3: Omnimodal world models for physical {AI}.
\newblock \emph{arXiv preprint arXiv:2606.02800}, 2026{\natexlab{b}}.

\bibitem[Chi et~al.(2025)Chi, Xu, Feng, Cousineau, Du, Burchfiel, Tedrake, and
  Song]{dp}
C.~Chi, Z.~Xu, S.~Feng, E.~Cousineau, Y.~Du, B.~Burchfiel, R.~Tedrake, and
  S.~Song.
\newblock Diffusion policy: Visuomotor policy learning via action diffusion.
\newblock \emph{The International Journal of Robotics Research}, 44\penalty0
  (10-11):\penalty0 1684--1704, 2025.

\bibitem[Zhao et~al.(2023)Zhao, Kumar, Levine, and Finn]{act}
T.~Z. Zhao, V.~Kumar, S.~Levine, and C.~Finn.
\newblock Learning fine-grained bimanual manipulation with low-cost hardware.
\newblock \emph{arXiv preprint arXiv:2304.13705}, 2023.

\end{thebibliography}
\clearpage
\renewcommand{\thesection}{\Alph{section}}
\setcounter{section}{1}
\section*{Supplementary Material}

This document provides five supplementary sections.
\S\ref{app:system} details the \ours class hierarchy and the per-step message-passing sequence between MuJoCo and Isaac Sim.
\S\ref{app:preprocessing} describes the offline asset preprocessing pipeline for physics collision geometries, USD rendering assets, and grasp synthesis.
\S\ref{app:baselines} gives the fine-tuning hyperparameters and embodiment adaptations for each evaluated baseline.
\S\ref{app:controller} describes the two whole-body locomotion controllers (AMO and SONIC) integrated in \ours.
\S\ref{app:extended} reports multi-task benchmarking results of the foundation models across all 16 tasks, followed by the complete task list.

\subsection{System Architecture Details}
\label{app:system}

\subsubsection{Main Class Entities}
\label{app:classes}

\begin{figure}[t]
    \centering
    \includegraphics[width=0.95\linewidth]{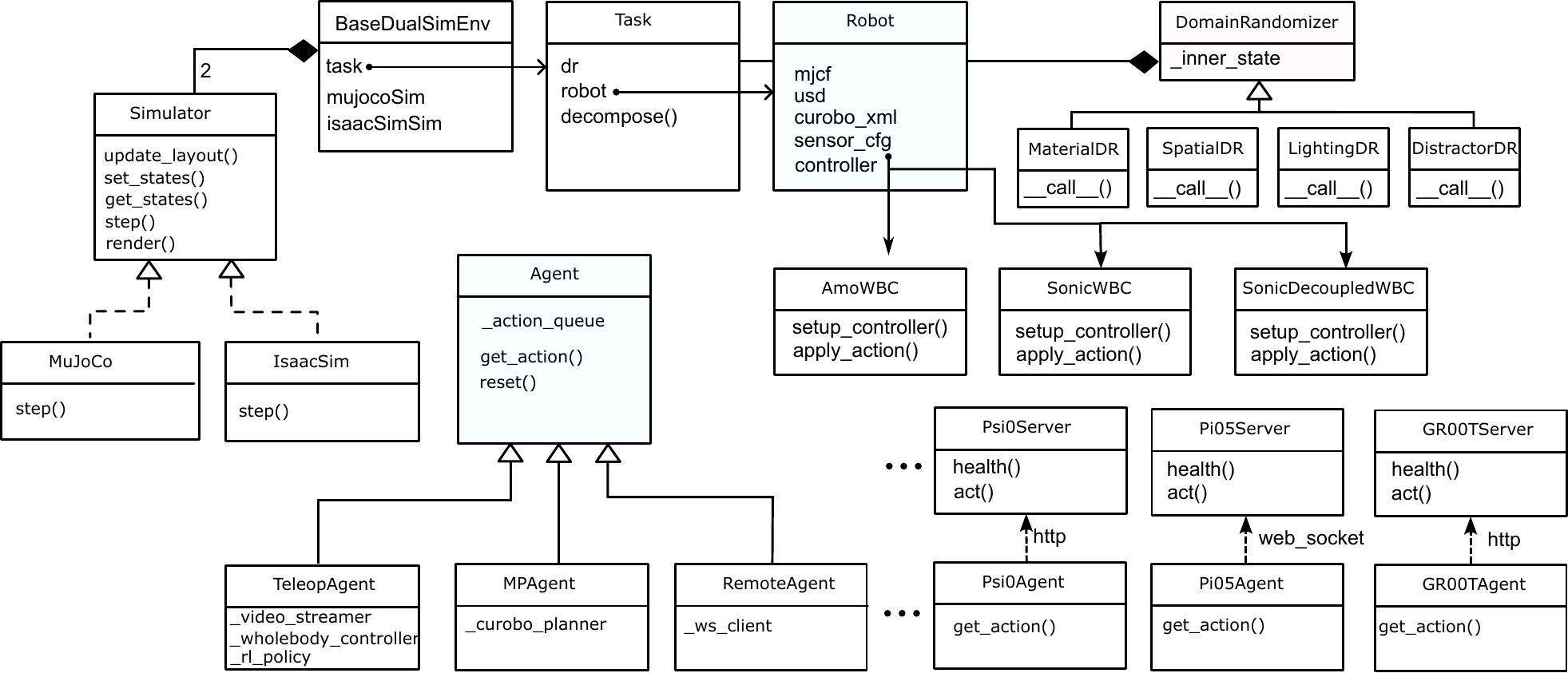}
    \caption{
    \textbf{Class Diagram of the \ours Framework.}
    \texttt{BaseDualSimEnv} owns two \texttt{Simulator} instances and a \texttt{Task}, and exposes a standard Gym interface.
    \texttt{Task} composes a \texttt{DomainRandomizer} and a \texttt{Robot} and drives scene randomization at each reset.
    The \texttt{Agent} hierarchy (teleoperation, motion planning, remote inference) shares a unified \texttt{get\_action} interface, while whole-body controllers (\texttt{AmoWBC}, \texttt{SonicWBC}, \texttt{SonicDecoupledWBC}) translate high-level commands into low-level joint targets.
    Policy servers (\texttt{Psi0Server}, \texttt{Pi05Server}, \texttt{GR00TServer}) run as independent processes and are accessed over HTTP or WebSocket.
    }
    \label{fig:class_diagram}
\end{figure}

Figure~\ref{fig:class_diagram} shows the class hierarchy of \ours; we describe each component below.

\paragraph{BaseDualSimEnv.}
\texttt{BaseDualSim} (\texttt{envs/base\_dual\_env.py}) is the central \texttt{gymnasium.Env} that holds references to both a \texttt{MujocoSimulator} and an \texttt{IsaacSimSimulator}, along with the active \texttt{Task}.
It initializes Isaac Sim lazily — the \texttt{SimulationApp} is created only when \texttt{"isaac"} appears in \texttt{sim\_mode} — so MuJoCo-only workflows incur no rendering overhead.
All concrete task environments (\textit{e.g.}, \texttt{LocoManipulationEnv}, \texttt{TabletopGraspEnv}) inherit from \texttt{BaseDualSim} and add task-specific \texttt{reset()} and \texttt{step()} logic on top.

\paragraph{Simulator.}
\texttt{Simulator} (\texttt{core/simulator.py}) is an abstract base class defining five operations: \texttt{update\_layout()}, \texttt{set\_states()}, \texttt{get\_states()}, \texttt{step()}, and \texttt{render()}.
\texttt{MujocoSimulator} (\texttt{engines/mujoco.py}) implements the physics backend: it runs the rigid-body contact simulation at 500\,Hz and manages robot actuation, collision geometries, and proprioceptive state readout.
\texttt{IsaacSimSimulator} (\texttt{engines/isaacsim.py}) implements the rendering backend: it consumes synchronized joint and pose states from MuJoCo and produces photorealistic ray-traced RGB frames via NVIDIA Isaac Sim's USD/Replicator pipeline.

\paragraph{Task.}
\texttt{Task} (\texttt{core/task.py}) is an abstract class that bundles all task-specific configuration: the \texttt{DRManager} instance, the \texttt{Robot} descriptor, sensor configs, and Gym action/observation spaces.
Its \texttt{reset()} method sequences through all registered randomizers — scene, target, container, distractor, spatial, lighting, camera, and material — building a \texttt{Layout} object that both simulators ingest to rebuild the scene.
Concrete subclasses (\textit{e.g.}, \texttt{G1WholebodyXMovePickTaskTeleop}) additionally implement \texttt{check\_success()}, a natural-language \texttt{instruction} property, and \texttt{decompose()} for automated subtask generation.

\paragraph{Robot.}
\texttt{Robot} (\texttt{core/robot.py}) is a configuration container that holds the MJCF physics model path, USD rendering model path, CuRobo kinematics config, sensor configurations, and controller config.
Concrete implementations (\textit{e.g.}, \texttt{G1Wholebody}, \texttt{G1Sonic}) extend \texttt{Robot} with mixin traits such as \texttt{HeadCamMountable}, \texttt{HasDexterousHand}, and \texttt{CuRoboMixin}, and declare robot-specific constants like joint names and default poses.
The \texttt{Robot} class does not execute control directly; it passes its asset paths to the \texttt{Simulator} backends and its kinematics config to the whole-body controller modules.

\begin{figure}[t]
    \centering
    \includegraphics[width=0.95\linewidth]{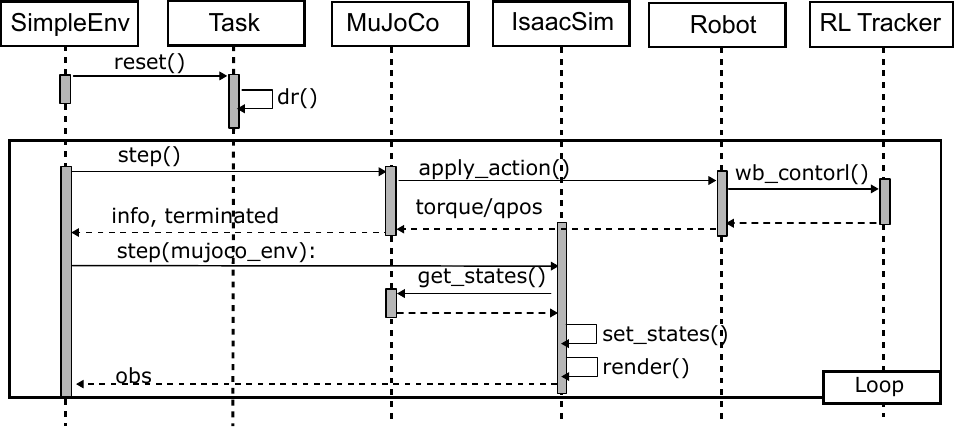}
    \caption{
    \textbf{Sequential Diagram of a Single \texttt{env.step()} Call.}
    Each invocation of \texttt{env.step(action)} (line~8 of Algorithm~1) triggers the following sequence:
    \texttt{MujocoSimulator} advances the physics loop at 500\,Hz and writes the updated robot and object states to a shared buffer;
    \texttt{IsaacSimSimulator} reads those states, renders photorealistic RGB frames at 50\,Hz, and returns the visual observation;
    \texttt{BaseDualSim} then queries the \texttt{Task} for the reward and termination condition before surfacing the full step output to the agent.
    }
    \label{fig:sequence_diagram}
\end{figure}
\vspace{-1em}

\paragraph{DomainRandomizer.}
\texttt{DRManager} (\texttt{dr/manager.py}) maintains a named registry of \texttt{Randomizer} instances and dispatches randomization calls at each \texttt{Task.reset()}.
Its \texttt{set\_level()} method adjusts active configurations to one of three difficulty levels: Level~0 adds distractor objects and background variation, Level~1 additionally randomizes material shaders and lighting, and Level~2 further varies object and robot spatial poses.
Four leaf randomizers each implement a \texttt{\_\_call\_\_()} interface: \texttt{MaterialDR} samples NVIDIA vMaterials surface shaders, \texttt{SpatialDR} varies object and robot initial poses, \texttt{LightingDR} randomizes light positions and intensities, and \texttt{DistractorDR} populates the scene with randomly chosen distractor objects.

\paragraph{Agent.}
\texttt{BaseAgent} (\texttt{agents/base\_agent.py}) is the abstract policy interface; all subclasses implement \texttt{get\_action(observation, instruction)} and manage an internal \texttt{\_action\_queue} that buffers predicted action chunks for step-by-step execution.
\texttt{MotionPlanAgent} (\texttt{agents/mp\_agent.py}) wraps a CuRobo planner: it converts the planned kinematic trajectory into a sequence of \texttt{ActionCmd} objects and replays them deterministically, enabling fully automated data collection without a human operator.
\texttt{PicoDecoupledAgent} (\texttt{agents/pico\_decoupled\_agent.py}) is the teleoperation agent: it streams egocentric stereo video to a PICO XR headset via \texttt{TCPVideoSender}, receives hand-pose commands retargeted through inverse kinematics, and routes the resulting upper-body targets through the decoupled WBC pipeline.

\paragraph{RemoteAgent.}
VLA and WAM policies run as independent server processes (\texttt{Psi0Server}, \texttt{Pi05Server}, \texttt{GR00TServer}), each exposing a \texttt{health()} readiness check and an \texttt{act()} inference endpoint over HTTP or WebSocket.
The corresponding client-side agents (\texttt{Psi0Agent}, \texttt{Pi05Agent}, \texttt{Gr00tN16Agent}) package RGB observations and proprioceptive state into the server's expected format, issue a blocking network call, and enqueue the returned action chunk into the shared \texttt{\_action\_queue}.
This server--client split allows policies to run on separate GPUs or machines, decoupling inference throughput from the simulation host's memory budget.

\paragraph{Whole-Body Controllers.}
\ours supports three whole-body controller (WBC) implementations, each following a common \texttt{setup\_controller()} / \texttt{apply\_action()} interface.
\texttt{AmoWBC} wraps the AMO controller~\cite{li2025amo}, which decouples upper- and lower-body control and uses a learned velocity-tracking policy for the base while accepting end-effector targets for the arms.
\texttt{SonicWBC} (\texttt{agents/sonic\_wbc\_agent.py}) and \texttt{SonicDecoupledWBC} (\texttt{agents/sonic\_decoupled\_wbc\_agent.py}) both build on the SONIC framework~\cite{luo2025sonic}: the former forwards raw whole-body motion-tracking commands via the Unitree SDK, while the latter instantiates the full \texttt{decoupled\_wbc} pipeline — including a G1 kinematic model and a WBC policy — to convert high-level joint targets into low-level position commands at 50\,Hz.

\subsubsection{Sequential Diagram}
\label{app:sequence}

Figure~\ref{fig:sequence_diagram} expands \texttt{env.step(action)} (line~8 of Algorithm~1) into its internal message-passing sequence.
Within each step, \texttt{MujocoSimulator} integrates physics at 500\,Hz and writes updated states to a shared buffer; \texttt{IsaacSimSimulator} reads those states and renders RGB frames at 50\,Hz; and \texttt{BaseDualSim} queries the \texttt{Task} for the reward and termination signal before returning the step output.
This producer--consumer ordering ensures every inference step receives a physically consistent, photorealistic observation while allowing MuJoCo to sub-step ahead of the slower rendering pass.

\subsection{Offline Object Preprocessing Pipeline}
\label{app:preprocessing}

\begin{figure}[t]
    \centering
    \includegraphics[width=0.95\linewidth]{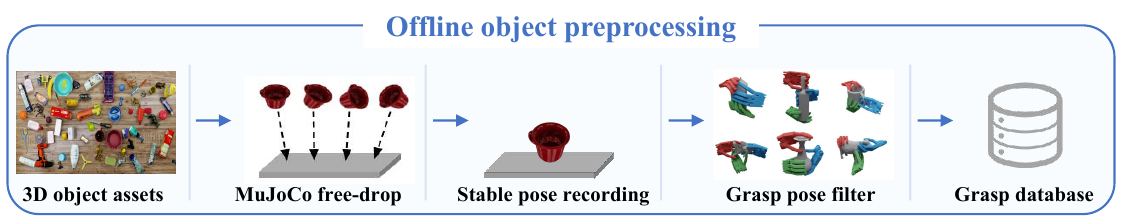}
    \caption{
    \textbf{Offline Object Preprocessing Pipeline.}
    Raw mesh assets from Objaverse and GraspNet-1B are processed for two purposes:
    (1) \textit{For Physics} — CoACD convex decomposition generates MuJoCo collision geometries, and stable resting poses are determined by dropping each object in simulation;
    (2) \textit{For Rendering} — meshes are converted to USD format with high-resolution PBR textures for Isaac Sim.
    BoDex \cite{chen2024bodex} synthesizes dexterous grasp poses on the stable-pose meshes; the results are cached alongside each asset for use at task initialization.
    }
    \label{fig:offline_object_preprocessing}
\end{figure}

Figure~\ref{fig:offline_object_preprocessing} shows the offline preprocessing pipeline that runs once per asset and produces all artifacts needed for physics, rendering, and grasp planning.

Assets are drawn from Objaverse~\cite{deitke2023objaverse} (1{,}500$+$ objects) and GraspNet-1B~\cite{fang2020graspnet} (75 objects).
After normalization, each mesh is processed in two parallel tracks:
the \textit{physics track} applies CoACD~\cite{wei2022coacd} convex decomposition to generate MuJoCo collision geometries and drops each object in simulation to enumerate stable resting poses;
the \textit{rendering track} converts meshes to USD format with high-resolution PBR textures for Isaac Sim.

BoDex~\cite{chen2024bodex} synthesizes dexterous grasp configurations for each stable pose offline, caching quality scores and approach depths per asset.
At task initialization, \ours samples a stable pose via \texttt{SpatialDR}, retrieves the cached grasps, and passes the selected target to CuRobo~\cite{curobo_v2} for trajectory generation — no online grasp synthesis is required at training or evaluation time.

\subsection{Baseline Implementation Details}
\label{app:baselines}

\textbf{$\Psi_0$}~\cite{wei2026psi0} is a 2.5B humanoid foundation model with a Qwen3-VL-2B vision backbone and a flow-matching diffusion transformer action expert.
We fine-tune the released pre-trained checkpoint for 40{,}000 steps on 8 NVIDIA A100 GPUs using DDP, with a per-device batch size of 16, a cosine-annealed learning rate of $1\times10^{-4}$, 1{,}000 warmup steps, and bf16 mixed precision.
The action head is configured with a 36-dimensional output and a chunk size of 30, matching the \ours whole-body action space.

\textbf{DreamZero}~\cite{dreamzero} extends the Wan2.1-I2V video generation model (14B parameters) to jointly predict future video frames and robot actions via a flow-matching DiT action head, enabling planning in image space during execution.
We initialize from the \textit{DreamZero-AgiBot} pretrained checkpoint and apply LoRA fine-tuning (rank~4, $\alpha$~4, targeting \texttt{q,k,v,o,ffn.0,ffn.2} layers) for 40{,}000 steps on eight NVIDIA A100 GPUs using DeepSpeed ZeRO-2, with a per-device batch size of 1, a cosine-annealed learning rate of $1\times10^{-4}$, and a 5\% linear warmup.
The action head outputs a 36-dimensional action at a step horizon of 48; video observations are rendered at $320\times176$ resolution and encoded by the Wan2.1 VAE, with the number of context frames set per task (17, 25, or 33 frames corresponding to chunk sizes of 2, 3, or 4, respectively).

\textbf{Cosmos3}~\cite{cosmos3} is a 16B Mixture-of-Transformers world foundation model that couples a Qwen3-VL-8B understanding branch with a generation expert and jointly denoises video latents and action tokens with a shared rectified-flow objective.
We initialize from the public Cosmos3-Nano checkpoint, register the G1 embodiment as a new domain with a 36-dimensional action space, and fully fine-tune the generation expert, timestep embedder, VAE projections, and newly initialized action projections while keeping the understanding branch frozen.
Training runs for 40{,}000 iterations on 32 NVIDIA H100 GPUs with a base learning rate of $2\times10^{-4}$ under a linear-decay schedule, a $5\times$ multiplier on the action projections, and $10\times$ weights on the video and action flow-matching losses.
Each sample is a 33-frame egocentric clip at 50\,fps encoded by the Wan2.2 VAE and paired with a 32-step chunk of normalized actions conditioned on the proprioceptive state.

\textbf{$\pi_{0.5}$}~\cite{intelligencepi05} shows strong generalization on mobile dual-arm platforms, but its released checkpoint supports only a 30-dimensional action space.
We expand the action dimension to 36 with a chunk size of 16, padding the weights of the affected linear layers to accommodate the larger space.
To bridge the embodiment gap between the original training distribution and humanoid morphologies, we raise the learning rate from $1\times10^{-5}$ to $1\times10^{-4}$ and the global batch size from 32 to 128, fine-tuning from the \textit{Pi05\_DROID} checkpoint ported to PyTorch.

\textbf{GR00T N1.6}~\cite{bjorck2025gr00t} is a 3B VLA pretrained for general robot manipulation.
We fine-tune from the released checkpoint for 20{,}000 steps on three NVIDIA A100 GPUs (global batch size 24, lr $1\times10^{-4}$ cosine), using all default hyperparameters.
As the RTC inference code is unavailable, we use a standard sequential scheme conditioning each prediction on the most recently executed observation.

\textbf{Diffusion Policy (DP)}~\cite{dp} uses a pre-trained ResNet-18 as the visual encoder, with a learning rate of $1\times10^{-4}$ and a global batch size of 32.
Training runs for 40{,}000 steps on two NVIDIA A100 GPUs, taking approximately 15 hours per task.
Despite fitting the training data reasonably, DP fails on most evaluation tasks; we attribute this to the insufficient visual capacity of the UNet-based architecture for the diverse visual conditions in \ours.
At inference time, we perform 100 denoising steps to recover actionable trajectories from noise.

\textbf{Action Chunking with Transformers (ACT)}~\cite{act} is reconfigured with a 36-dimensional action head and a chunk size of 100, using a transformer comprising 4 encoder layers and 1 decoder layer following the publicly released LeRobot configuration~\cite{cadene2026lerobot}.
All other hyperparameters are kept identical to those used for DP.

\begin{table*}[t]
\centering
\small
\caption{Representative tasks in our benchmark.}
\label{tab:teleop_tasks}

\renewcommand{\arraystretch}{1.05}
\setlength{\tabcolsep}{4pt}

\newcolumntype{T}{>{\ttfamily\small}p{5.0cm}}

\emergencystretch=6pt
\begin{tabularx}{\textwidth}{T p{2.0cm} p{1.0cm} X}
\toprule
\textbf{Task Name} & \textbf{Category} & \textbf{Teleop} & \textbf{Success Criteria} \\
\midrule

BendPick
& Rigid & Yes
& Lift the target object more than 5 cm from the table. \\

BendPickAndPlace
& Rigid & Yes
& Place the target object into the container. \\

BendHandover
& Bimanual & Yes
& Handover the object and place it into the container. \\

Handover
& Bimanual & Yes
& Handover the object and place it into the container. \\

OpenTrashCan
& Articulated & Yes
& Open the trash can lid beyond the target angle threshold. \\

PushOfficeChair
& Non-prehensile & Yes
& Push the office chair beyond the target distance. \\

OpenFaucet
& Articulated & Yes
& Rotate the faucet handle beyond the target angle. \\

OpenOven
& Articulated & Yes
& Open the oven door beyond the target angle threshold. \\

CloseDoor
& Articulated & Yes
& Close the door to the target closed state. \\

XMovePick
& Mobile & Yes
& Move laterally and lift the target object. \\

XMoveBendPick
& Mobile & Yes
& Move laterally, bend down, and lift the target object. \\

LocomotionPickBetweenTables
& Whole-body & Yes
& Transfer the object from one table to another table container. \\

PickAndPlaceAndHugContainer
& Whole-body & Yes
& Place two objects into the container, lift it, and move it onto another table. \\

BendPickMP
& Rigid & No
& Lift the target object more than 5 cm from the table. \\

TableTopGraspMp
& Rigid & No
& Lift the target object more than 5 cm from the table. \\

LocoPickBetweenTablesMP
& Whole-body & No
& Transfer the object from one table to another table. \\

\bottomrule
\end{tabularx}
\end{table*}

\subsection{Whole-Body Controller Details}
\label{app:controller}

\ours integrates two whole-body locomotion controllers that share a common teleoperation interface but differ in their internal control architecture.
Both controllers receive the same high-level inputs — upper-body joint targets from the teleoperation or policy layer and a 4-dimensional navigation command $(v_x, v_y, v_{\mathrm{yaw}}, q_{\mathrm{yaw}})$ — and output low-level joint position commands to the Unitree G1 at 500\,Hz via MuJoCo.

\textbf{AMO}~\cite{li2025amo} adopts a decoupled architecture that treats upper-body and lower-body control as two independent problems.
The lower body is governed by a learned RL locomotion policy (\texttt{AMO\_Policy} in \texttt{robots/policy/AMO\_Policy.py}) that takes IMU measurements and proprioceptive joint states as input and outputs 15 lower-body joint targets (hip, knee, ankle, and waist joints) at high frequency.
The upper body is controlled separately by directly tracking the joint targets received from the high-level policy or teleoperator, so arm and hand movements do not feed back into the locomotion policy, giving the operator more predictable upper-body control at the cost of tighter coupling between arm posture and balance.

\textbf{SONIC}~\cite{luo2025sonic} takes an end-to-end whole-body motion tracking approach.
\ours provides two SONIC-based implementations: \texttt{SonicWbcAgent} (\texttt{agents/sonic\_wbc\_agent.py}), which forwards whole-body commands directly through the Unitree SDK bridge (\texttt{gear\_sonic}), and \texttt{SonicDecoupledWbcAgent} (\texttt{agents/sonic\_decoupled\_wbc\_agent.py}), which runs the full \texttt{decoupled\_wbc} pipeline internally — instantiating a G1 kinematic model and a WBC policy — to convert upper-body joint targets and navigation commands into full-body position commands at 50\,Hz.
Because SONIC tracks the operator's whole-body motion as a unified trajectory rather than handling each segment independently, it tends to produce smoother and more natural whole-body demonstrations, particularly for tasks requiring coordinated arm--leg movements such as mobile pick-and-place.

\subsection{Extended Benchmark Results}
\label{app:extended}

Beyond the single-task fine-tuning results in Table~1 of the main paper, we benchmark the foundation models by jointly fine-tuning each on all 16 tasks and report per-task success rates under the three domain-randomization levels (Level~0 / Level~1 / Level~2) in Table~\ref{tab:benchmark_all}.
ACT and DP are excluded from this setting since they are designed for single-task training.
Each task is evaluated with 20 trials per level to tighten the confidence intervals.

\begin{table}[h]
\centering
\footnotesize
\setlength{\tabcolsep}{4pt}
\resizebox{\textwidth}{!}{%
\begin{tabular}{@{} l c c c c c @{}}
\thickrule
\tablesp
\textbf{Task} & \textbf{$\Psi_0$~\cite{wei2026psi0}} & \textbf{$\pi_{0.5}$~\cite{intelligencepi05}} & \textbf{GR00T N1.6~\cite{bjorck2025gr00t}} & \textbf{Cosmos3-Nano~\cite{cosmos3}} & \textbf{DreamZero~\cite{dreamzero}} \\
\tablesp
\thinrule
\tablesp
\texttt{BendHandoverTeleop} & 75/50/60 & 65/40/30 & 80/80/0 & \ul{80/75/65} & \textbf{70/80/90} \\
\texttt{BendPickAndPlaceTeleop} & 10/20/0 & 15/10/5 & 20/10/0 & \textbf{15/25/20} & \textbf{30/20/10} \\
\texttt{BendPickMP} & \textbf{100/100/100} & 5/0/0 & 45/30/40 & 0/0/0 & \ul{90/90/80} \\
\texttt{BendPickTeleop} & 100/50/50 & 75/75/70 & \textbf{100/100/80} & 70/65/35 & 80/80/60 \\
\texttt{CloseDoorTeleop} & 95/95/75 & 90/75/95 & 100/95/100 & 100/100/95 & \textbf{100/100/100} \\
\texttt{HandoverTeleop} & \textbf{70/70/100} & 55/35/45 & 65/65/0 & \ul{65/100/70} & 70/50/30 \\
\texttt{LocomotionPickBetweenTablesTeleop} & \textbf{70/50/60} & 35/15/20 & 55/50/10 & \ul{70/50/20} & 50/30/30 \\
\texttt{OpenFaucetTeleop} & \textbf{10/10/35} & 5/0/0 & 0/0/0 & \ul{15/5/10} & 10/0/0 \\
\texttt{OpenOvenTeleop} & 65/25/10 & 55/25/50 & \ul{85/60/90} & 65/60/75 & \textbf{90/80/70} \\
\texttt{OpenTrashCanTeleop} & \ul{45/55/25} & 5/5/5 & \textbf{65/50/75} & 20/10/15 & 20/10/10 \\
\texttt{PickAndBendPlaceMP} & 20/55/40 & 60/45/20 & \textbf{90/75/85} & \ul{85/75/75} & 80/70/50 \\
\texttt{PickAndPlaceAndHugContainerTeleop} & 25/25/10 & 35/15/35 & \textbf{80/80/20} & \ul{40/50/30} & 50/50/10 \\
\texttt{PushOfficeChairTeleop} & \ul{100/100/65} & 95/90/60 & \textbf{100/100/100} & 100/90/60 & 100/70/40 \\
\texttt{TabletopGraspMP} & \textbf{100/100/80} & 40/10/25 & 75/90/60 & 30/30/25 & \ul{90/100/70} \\
\texttt{TabletopHandoverMP} & \ul{45/50/10} & 0/5/5 & 0/5/15 & \textbf{95/55/60} & 0/0/0 \\
\texttt{XMovePickTeleop} & 100/100/60 & 15/15/25 & 0/0/0 & \ul{95/95/75} & \textbf{100/100/100} \\
\tablesp
\thinrule
\tablesp
\textbf{Average} & \textbf{64.4/59.7/48.8} & 40.6/28.8/30.6 & 60.0/55.6/42.2 & 59.1/55.3/45.6 & \ul{64.4/58.1/46.9} \\
\tablesp
\thickrule
\end{tabular}%
}
\caption{
\textbf{Multi-Task Benchmarking Results.}
Success rates (\%) under three domain-randomization levels (\textit{Level 0 / Level 1 / Level 2}) of foundation models jointly fine-tuned on all 16 tasks.
Each entry averages 20 trials per task and level (10 for DreamZero due to time constraints).
}
\label{tab:benchmark_all}
\end{table}

Averaged over tasks, $\Psi_0$~\cite{wei2026psi0} leads the benchmark, followed closely by DreamZero, GR00T N1.6, and Cosmos3-Nano~\cite{cosmos3}, whereas $\pi_{0.5}$ lags behind; success rates generally decrease with the domain-randomization level for all models.
Tasks with unambiguous geometric success signals (\texttt{CloseDoor}, \texttt{PushOfficeChair}) are solved by nearly all models, whereas fine-grained contact tasks (\texttt{OpenFaucet}) stay at or below 35\% for every model, confirming a meaningful difficulty gradient.
Every task is solved at a non-trivial rate by at least one model, establishing these results as reference baselines for future comparisons.

\end{document}